\pdfoutput=1

\documentclass[11pt]{article}

\usepackage[final]{acl}

\usepackage{times}
\usepackage{latexsym}
\usepackage{amsmath}
\usepackage{amsthm}
\usepackage{graphicx}
\usepackage{hyperref}

\usepackage[T1]{fontenc}
\usepackage[utf8]{inputenc}
\usepackage{microtype}
\usepackage{inconsolata}

\usepackage{algorithm}

\usepackage{algpseudocode}

\usepackage{subcaption}

\usepackage{amssymb}
\usepackage{comment}
\newcommand{\hidecomment}[1]{}

\newtheorem{thm}{Theorem}[section]
\newtheorem{prop}[thm]{Proposition}

\newtheorem{defn}[thm]{Definition}

\def\R{{\mathbb R}}

\def\xi{{\cal X}}

\DeclareMathOperator*{\argmax}{arg\,max}

\usepackage{pdflscape}

\usepackage[T1]{fontenc}

\usepackage[utf8]{inputenc}

\usepackage{microtype}

\usepackage{inconsolata}

\usepackage{url}
\usepackage{arydshln}

\usepackage{multirow}

\newcommand{\paragraphown}[1]{

\textbf{#1}
}

\title{Divergent Token Metrics: Measuring degradation to \\ 
prune away LLM components -- and optimize quantization \vspace{.2cm}}

 \author{\hspace{-.4cm}Björn Deiseroth$^{1,2,3}$
 \And\hspace{-.75cm} Max Meuer$^1$
         \And\hspace{-1.5cm} Nikolas Gritsch$^{1,5}$
         \And\hspace{-.35cm} Constantin Eichenberg$^1$\\
         \hspace{-12.7cm}\texttt{\{bjoern.deiseroth, max.meuer, nikolas.gritsch, constantin.eichenberg\}@aleph-alpha.com}
         \AND\vspace{-10cm}\hspace{-.445cm} Patrick Schramowski$^{2,3,4}$\hspace{.6cm}
         \And Matthias Aßenmacher$^{5,6}$
         \And \hspace{.25cm}Kristian Kersting$^{2,3,4}$\\
         \hspace{-10.45cm}\texttt{patrick.schramowski@dfki.de  \  
         matthias@stat.uni-muenchen.de \ 
         kersting@cs.tu-darmstadt.de} \vspace{.5cm}
         \\
         \hspace{-10.75cm} $^1$ Aleph Alpha @ IPAI \hspace{.35cm} $^2$ Technical University Darmstadt\\
         \hspace{-10.75cm} $^3$ Hessian Center for Artificial Intelligence (hessian.AI) \\
         \hspace{-10.75cm} $^4$ German Center for Artificial Intelligence (DFKI) \\
         \hspace{-10.75cm} $^5$ LMU Munich \hspace{.35cm} $^6$ Munich Center for Machine Learning (MCML)\\  
         \\ 
         }

\begin{document}
\maketitle

\def\thefootnote{}\footnotetext{
\hspace{-15px}\url{
https://github.com/Aleph-Alpha/Divergent_Tokens}}

\begin{abstract}
Large Language Models (LLMs) have reshaped natural language processing with their impressive capabilities. 
However, their ever-increasing size has raised concerns about their effective deployment and the need for LLM compression. 
This study introduces the \textit{Divergent Token} Metrics (DTMs), a novel approach to assessing compressed LLMs, addressing the limitations of traditional perplexity or accuracy measures that fail to accurately reflect text generation quality. DTMs measure token divergences that allow deeper insights into the subtleties of model compression, in particular, when evaluating components' impacts individually.
Utilizing the \emph{First Divergent Token} Metric (FDTM) in model sparsification reveals
that 25\% of all attention components can be pruned beyond 90\% on the Llama-2 model family, still keeping SOTA performance. For quantization, FDTM suggests that more than 80\% of the parameters can be naively transformed to int8 without special outlier management.
These evaluations indicate the necessity of choosing appropriate compressions for parameters individually---and that FDTM can identify those---while standard metrics result in deteriorated outcomes.

\end{abstract}


\section{Introduction}
\label{sec:intro}

\begin{figure*}[t]
\centering
    \begin{subfigure}{.8\linewidth}
\includegraphics[width=\linewidth]{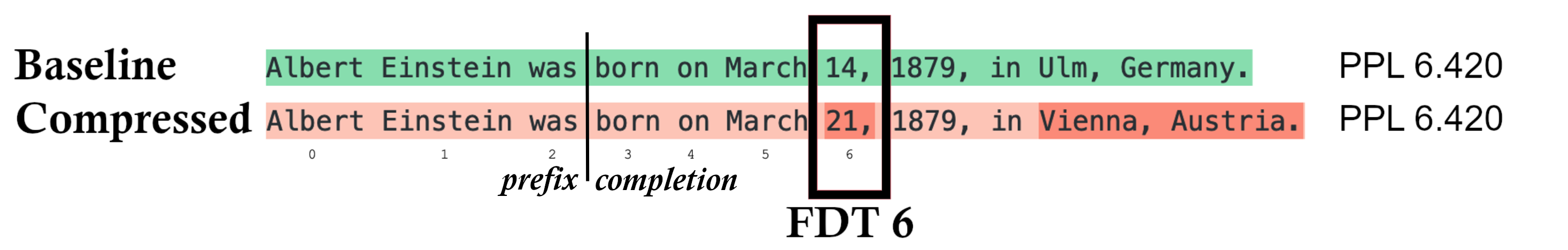}
    \end{subfigure}
\caption{Illustration of a diverging generation process. Given the 3-token prefix as prompt, a baseline and its compressed model generate 8 subsequent tokens. Our proposed metric points to the first divergent token (FDT). The FDT may cause further divergence during the iterative generation process. Note how both models score the same perplexity value, as it does not reflect the actual sampling process (\textit{cf.} Fig.~\ref{fig:sparse_random}, Sec.~\ref{sec:compression} for an empirical exploration).}
\label{fig:metric-example}
\end{figure*}

Cutting-edge Large Language Models (LLMs) based on the transformer architecture \citep{vaswani2017attention} have revolutionized Natural Language Processing with their exceptional performance, notably exemplified by the GPT series~\citep{radford2018improving,radford2019language,brown2020language,bubeck2023sparks} in text generation. 
However, these models have grown massively, even exceeding half a trillion parameters~\cite{chowdhery2022palm}. Although the large number of parameters aid in early training convergence, their practical utility and true necessity remain unclear. 
In particular, for the attention mechanism, it was hinted that after some training convergence, certain heads dominate the inference process~\cite{michel2019sixteen}. 

Compression strategies such as sparsification and quantization can enhance the efficiency of the parameters.
Current metrics, however, either average too coarsely, such as perplexity, or are by design too specific, such as standard NLP benchmarks.
Both fail to capture the diverging performance nuances introduced early on by the compression as they ignore the actual discontinuous text generation process. This, however, is the main application of the final model, and so we argue that they are therefore insufficient measures for the performance of the compressed model.
This misalignment can lead to unwanted subtle discrepancies in generation, 
such as grammatical errors or a mismatch in numbers, that cause subsequent divergences. As we will show, this even occurs when overall metrics, such as perplexity, appear satisfactory (\textit{cf.} Prop.~\ref{prop:ppl_discontinuity}, Sec.~\ref{sec:compression}, Tab.~\ref{tab:comp_model_benchmarks}). 

To meet these challenges, we introduce the family of \textit{Divergent Token} Metrics (DTMs) in Sec.~\ref{sec:metrics}. 
These metrics are tailored to measure the \emph{model divergence} of LLMs throughout the compression process
and in relation to the actual generation procedure on a token basis, as shown in Fig.~\ref{fig:metric-example}.
We demonstrate that the \textit{First Divergent Token} Metric (FDTM) and the \textit{Share of Divergent Tokens} Metric (SDTM) offer a more nuanced evaluation compared to perplexity. They also enable individual component evaluation to rank parts of the model best suited for compression, thus enabling meaningful compression while preserving text generation quality.
Based on FDT probing, we introduce new strategies for sparsifying and quantizing models in Sec.~\ref{sec:compression}.

Specifically, our proposed individual component sparsification indicates significant differences 
in component utilization across layers. 
For the first time, 
we show that 25\% percent of the models' attention components can be pruned beyond 90\%, and several even entirely removed, while preserving a  single-digit perplexity. 
Consequently, a sparse matrix format can be employed to accelerate computational efficiency.
Likewise, for precision reduction, we show that sorting components by FDTM coincidentally correlates to sorting by their induced number of outliers when being naively converted to int8. FDTM identifies the optimal 80\% of the components that maintain overall performance without specific outlier handling.
The observed decline in performance with more outliers, and the significant influence of specific components on those, suggests reevaluating the applied normalization methods throughout the model.
We demonstrate that this level of precision goes beyond what standard perplexity and conventional NLP benchmarks can achieve.
The proposed Divergent Token Metrics closely reflect the generation process and so can be a measure to foster confidence in the deployed compressed models.

\section{Compression Principles}
\label{sec:background}
Model compression aims to reduce the hardware resources needed to operate the model.
However, doing so may sacrifice the accuracy of the model. 
To keep the loss as small as possible, 
a corrective measure is typically used.
Here, we discuss the most commonly used concepts and state-of-the-art methods for LLM sparsification and quantization.

\paragraphown{Outlier and Hessians.}
Most model compression methods rely either on the separation of outliers~\cite{dettmers2022llmint8,sun2023wanda} or the computation of a Hessian matrix~\cite{frantar2023gptq,frantar2023sparsegpt}.
Outliers usually refer to significantly larger values in magnitude occurring either in the weight matrix directly or in the activations during a forward pass.
As most computations are linear matrix multiplications, such outliers strongly influence the remaining entropy contained in consecutive computations. In the case of sparsification, outliers should be left intact, and values with the least magnitude, which are consequently the least influential, should instead be masked~\cite{han2015learning}.
On the other hand, Hessian matrices can be applied to correct errors~\cite{frantar2023gptq}. They can effectively be approximated by computing backpropagation gradients for a small number of samples and represent a second-order approximation to reconstruct the original model. 

\paragraphown{Sparsification.} 
The goal of sparsification is a reduction in the total number of weights and therefore a distillation of the relevant computation. Typically, this method is divided into ``structured'' and ``unstructured'' pruning. \textit{Structured-pruning} aims to locate dynamics, such as the irrelevance of an entire layer or dimension for a given use case, and prunes these entirely.
\textit{Unstructured-pruning} usually refers to weight masking, that is, setting irrelevant weights to 0. High levels of sparse matrix computations could result in more efficient kernels and computations. 
In scenarios where masks exceed a 90\% threshold, the implementation of a specialized sparse matrix format becomes feasible. This format predominantly stores the indices of non-zero weights. Although some additional storage is required for these indices, the overall requirement is reduced due to the exclusion of zero values. Moreover, this approach substantially improves computational performance.

\textit{Magnitude pruning} selects the masking of weights based only on their magnitudes. 
This is fast to compute but significantly degrades model performance when pruning large amounts simultaneously.
To resolve this issue, \textit{wanda}~\cite{sun2023wanda} proposes to sample a small amount of data and incorporates activations of the forward pass. This was shown to generate more effective one-shot pruning masks. 
 \textit{SparseGPT}~\cite{frantar2023sparsegpt} computes iterative Hessian approximations to select the lowest impact weights and correct the remaining.

Note that the incorporation of activations can, to some extent, be interpreted as a form of training. Moreover, despite these efforts, one-shot pruning has not yet produced directly usable models without further final fine-tunings. This is in particular the case for the high sparsity levels beyond 70\% that we target. Finally, 
there has not yet been any investigation of the individual components.

\paragraphown{Quantization.}
Model quantization refers to the reduction of the precision of the used numerical format. Usually, LLMs are trained in 16-bit floating point (fp16) and converted to 8-bit integer (int8) representations. 
The naive conversion of float matrices to integers is  \textit{AbsMax} rounding. This divides a number by the maximum value occurring in the matrix and multiplies by the largest available integer, as such, it spans a uniform representation grid. The largest float value is stored and multiplied for dequantization.
The most prominent methods for minimizing the introduced rounding errors are LLM.int8 and GPTQ.

\citet{dettmers2022llmint8} introduced \textit{LLM.int8()}, which identifies vectors containing outliers and retains them in their original fp16 form during the matrix multiplication of a forward pass. The vectors lacking outliers are fully quantized. The int8 weights and activations during the forward pass are subsequently multiplied and dequantized afterward. 
This allows them to be integrated with the fp16 representation of the outliers. Through empirical investigation
optimizing the trade-off between degradation in perplexity and the number of outliers preserved in fp16, 
they fixed an absolute outlier threshold. 

The \textit{GPTQ} framework offers a more robust quantization approach, in particular, to different integer bit precisions. It does not rely on any outlier detection mechanism or mixed precision computations---matrix multiplications with the weights are fully performed using integers. 
\citet{frantar2023gptq} introduce an efficient Hessian approximation and iteratively quantize the weights of the matrices while performing error corrections on the remaining weights.



\section{Model Divergence Metrics}
\label{sec:metrics}

Perplexity fails to identify minor variations in model degradation at an early stage.
This behavior is depicted in Fig.~\ref{fig:metric-example} and \ref{fig:sparse_random} and discussed in Sec.~\ref{subsec:tokenvsperplexity} and \ref{sec:compression} in more detail.
To assess model divergence and enhance the model compression process, we introduce token-based metrics specifically designed to detect those nuances occurring in early compression stages. We start by establishing our notation and presenting the perplexity metric (PPL). Subsequently, we introduce an enhanced variant of PPL and propose the \textit{Share of Divergent Tokens}  Metric (SDTM) and \textit{First Divergent Token} Metric (FDTM). We conclude by discussing the advantages of each metric compared to traditional measures based on perplexity when assessing the degradation of the generative performance of compressed models.




\subsection{Basic notation}
Let $F$ denote an auto-regressive model over a vocabulary \mbox{$\mathcal{V} = \{0,1,...,|\mathcal{V}|-1 \}$}, \mbox{$y = (y_1,..,y_N) \in \mathcal{V}^N$} an arbitrary token sequence and \mbox{$F(y) = F(y)_{ij} \in \R^{N \times |\mathcal{V}|}$} the model logits, with $i$ denoting the sequence and $j$ the respective vocabulary positions.
Given a prefix length $n < N$, we denote the token prefix $y_{:n} = (y_1,...,y_{n})$ and the greedily decoded completion up to index $N$ by $\mathcal{G}(F,y_{:n},N)$. It is defined recursively as follows: 
$\mathcal{G}(F,y_{:n},N)_{:n} = y_{:n}$, and for $n \leq i \leq N-1$ 
\begin{align*}
    \textstyle \mathcal{G}(F,y_{:n},&N)_{i+1}  \\
    &\textstyle = \argmax_{j} F(\mathcal{G}(F,y_{:n},N)_{:i})_{ij}. 
\end{align*}

\subsection{Perplexity (PPL)}
Given a ground truth sequence $y$ and model $F$, the negative log-likelihood of $y$ given $F$ is
\begin{align*}
    \textstyle \operatorname{NLL}(y,F,&n) \\
    &\textstyle = - \frac{1}{N-n}  \sum_{i=n}^{N-1} \log \mathbb{P} (y_{i+1} | y_{i},..,y_1), \notag
\end{align*}
with $\mathbb{P} (y_{i+1} | y_{i},..,y_1) = (\mathrm{softmax} \ F(y))_{iy_{i+1}}$. Then the \emph{perplexity (PPL)} is given by 
\begin{align*}
\operatorname{PPL}(y,F,n) = \exp(\operatorname{NLL}(y,F,n)).
\end{align*} 
A common practice in the literature, e.g.~\cite{dettmers2022llmint8}, is to measure model degradation as the increase in average perplexity over a given test dataset $\mathcal{D}$, e.g. randomly sampled from C4~\cite{c4dataset}. Usually, this metric is computed disregarding the prefix, i.e., with $\operatorname{PPL}(y,F) := \operatorname{PPL}(y,F,1)$.

\begin{figure}
\centering
\includegraphics[width=.75\linewidth,height=105px]{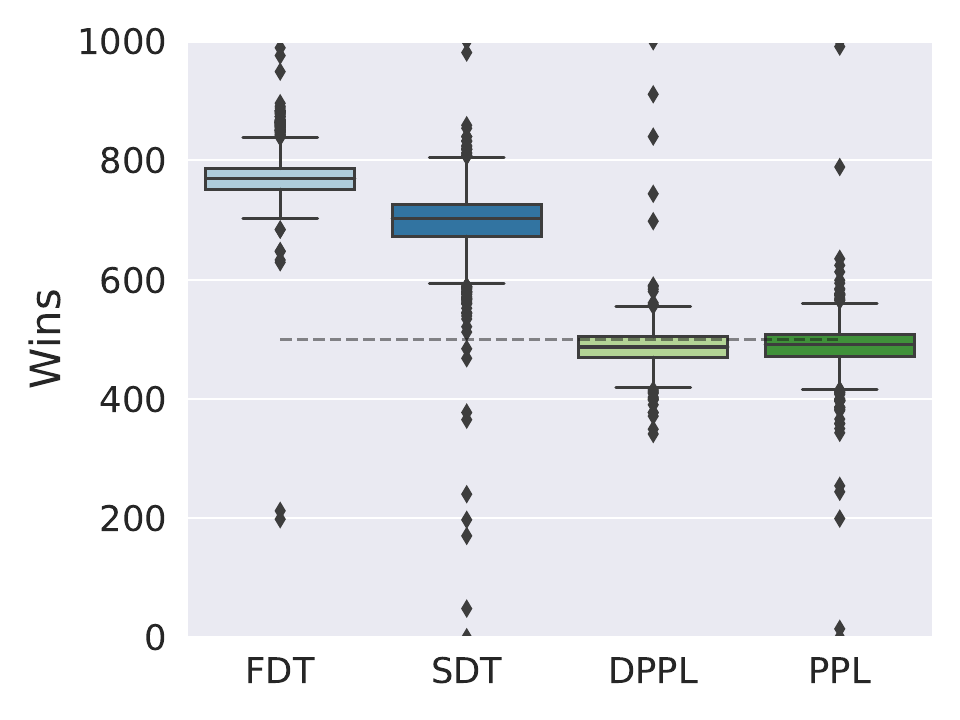}
\caption{Pruning lowest weights, and random weights. FDT is able to discriminate the cases. PPL exactly performs on the level of guessing. \textit{Cf.} Sec.~\ref{sec:sparse}.
\label{fig:sparse_random}}
\end{figure}


\subsection{Context aware model comparison}
First, we argue that standard evaluation does not reflect the typical generative model usage, i.e., there are no empty prompts, and as such, those positions should not be taken into account when evaluating the generative performance.
Moreover, when comparing a compressed model $F'$ to the original model $F$, one is interested to what extent \emph{the original behavior is kept}.  Therefore, we propose to use the outputs of the original model $F$ as a ground truth to assess the performance of the compressed model $F'$. This leads to the definition of the \emph{divergent perplexity (DPPL) metric} as
\begin{align}
    M_{\mathrm{DPPL}}(F,&F',y_{:n}, N) \label{e:ppl_div} \\
    &= \operatorname{PPL}(\mathcal{G}(F,y_{:n}, N),F',n)\;. \notag
\end{align}
Finally, let $\mathcal{D}$ be an arbitrary test dataset containing documents of potentially varying length. For a fixed  prompt length $n$ and completion length $N$, we define the  \emph{aggregated divergent perplexity metric} as
 the complete evaluation on the dataset:
\begin{align}
    \textstyle\mathcal{M}_{\mathrm{DPPL}}(&F,F',n,N) = \label{e:ppl_aggregated}\\
    & \textstyle \frac{1}{|\mathcal{D}|} \sum_{y \in \mathcal{D}} M_{\mathrm{DPPL}}(F,F',y_{:n},N). \notag
\end{align}

 DPPL already substantially improves discriminative capabilities over PPL, as we will demonstrate in the empirical evaluation.

\subsection{Divergent Token Metrics}
\paragraphown{SDT.}
To further improve on the expressiveness and interpretability of model divergence,
we propose the \emph{share of divergent tokens (SDT)} as follows:
\begin{align*}
    \textstyle \operatorname{SDT}(y,&F,n) \\
    &\textstyle = | \{i \geq n \colon \ \argmax_{j} \ F(y)_{ij} \neq y_{i+1} \} |, \notag
\end{align*}
$\operatorname{SDT}(y,F,n)$ can be interpreted as the number of times the model would need to be corrected during decoding to match the ground truth after consuming the prefix. 
This measure provides a more direct interpretation of the errors that occur during actual token generation, as opposed to estimating prediction certainties as PPL does.

\paragraphown{FDT.}
 In addition to SDT, we introduce the \textit{first divergent token (FDT)} as
\begin{align}
&{\textstyle\operatorname{FDT}(y,F,n)} \label{e:fdt} \\
&{\textstyle= \min \{i \geq n \colon  \argmax_{j} \ F(y)_{i,j} \neq y_{i+1} \}-n,   }\notag
\end{align}
with the convention that the minimum is equal to $N$ if the set on the right-hand side is empty.
Analogously to Eq.~\ref{e:ppl_div} and Eq.~\ref{e:ppl_aggregated}, we define $M_{\mathrm{SDT}}, M_{\mathrm{FDT}}$, $\mathcal{M}_{\mathrm{SDT}}$ and $\mathcal{M}_{\mathrm{FDT}}$ in the same way.


As an illustrative example, consider 
 computing $M_{\mathrm{FDT}}(F,F',y_{:n},N)$. We first perform a greedy decoding of $N-n$ tokens with the base model $F$ given the prefix $y_{:n}$. We then feed the sequence $\mathcal{G}(F,y_{:n}, N)$ into the compressed model $F'$ and find the first index greater than or equal to $n$, where the logit argmax of $F'$ differs from what $F$ generated. This computation can be done in a single forward pass similar to perplexity, and so is more efficient than accuracy-based evaluations.
Trivially, $0 \leq M_{\mathrm{FDT}}(F,F',y_{:n},N) \leq N - n$, where the upper bound is reached if and only if $F$ and $F'$ would generate the exact same sequence up to position $N$ given the prefix $y_{:n}$. 
Further note that $M_{\mathrm{FDT}}$ is symmetric, i.e. $M_{\mathrm{FDT}}(F,F',y_{:n},N) = M_{\mathrm{FDT}}(F',F,y_{:n},N)$, in contrast to PPL. 


In the following, we will ease the notation and omit $\mathcal{M}$, some parameters, or the words aggregated and metric, when they are clear from the context.

\subsection{Token vs. Perplexity Metrics}
\label{subsec:tokenvsperplexity}
It turns out that divergent token metrics offer a superior criterion for analyzing model performance degradation compared to perplexity-based metrics, especially in the context of greedy decoding. 
The main reason   
is that the greedy decoding operation $\mathcal{G}$ is a discontinuous function of the logits. To formalize this, let us discard the model itself and focus notation solely on the concept of logits.
\begin{defn}
    The operators and metrics from previous sections defined for models $F, F'$ are defined for logits $l,l' \in \R^{N \times |\mathcal{V}|}$ by replacing all occurrences of $F, F'$ with $l,l'$.
\end{defn}
\noindent
For example, $\mathcal{G}(l,y_{:n},N)_{i+1} = \argmax_j l_{ij}$, for $n \leq i \leq N$.



\begin{prop}\label{prop:ppl_discontinuity}
    Given any $y$, $N$ and $\varepsilon > 0$ there exist logits $l, l' \in \R^{N \times |\mathcal{V}|}$ such that
    \begin{align*}
        \textstyle |\operatorname{PPL}(y,l,1) - \operatorname{PPL}(y,l',1)| &< \varepsilon, \\
        \textstyle M_{\mathrm{SDT}}(l,l',y_{:1}, N) &= N.
    \end{align*}
\end{prop}
\begin{proof}
See App.~\ref{sec:proofs}.
\end{proof}
This means that even if the average perplexity of a compressed model matches the perplexity of the original model, the compressed model can produce a very different (and potentially worse) output when performing greedy decoding. Hence leading to a false positive. In practice, this is a severe issue since even a single diverging token can lead to a completely different subsequent output. 
It is illustrated in Fig.~\ref{fig:metric-example} and \ref{fig:sparse_random} and discussed in Sec.~\ref{sec:sparse}.

As described above, another option is to compute the perplexity with respect to the generated completions of the original model. This metric relates more reasonably to the share of divergent tokens (SDT):
\begin{prop} \label{prop:upper_bound}
The following upper bound holds:
    \begin{align*}
        {\textstyle M_{\mathrm{SDT}}(l,l',y_{:n}, N) \leq \frac{N-n}{\log 2} \log M_{\mathrm{DPPL}}(l,l',y_{:n}, N).}
    \end{align*}
\end{prop}
\begin{proof}
See App.~\ref{sec:proofs}.
\end{proof}
However, a comparable lower bound generally does not hold. In fact, in the case $l = l'$ we trivially have $M_{\mathrm{SDT}}(l,l,y_{:n}, N) = 0$. Further, the value of $M_{\mathrm{DPPL}}(l,l,y_{:n}, N)$ can still be as high as the maximal value, which occurs when $l$ is a perfectly flat distribution at any sequence index. This could lead to a false negative signal for the generation process.  

In conclusion, perplexity-based metrics suffer from false positives or false negatives when evaluating the degradation of generative performance. The case for FDT and SDT is quite straightforward in that they both directly measure the difference between model outputs in a \textit{what-you-see-is-what-you-get} manner.

Note that additional token-based metrics, such as the measurement of the distance between erroneous predictions, can be readily formulated. 
These metrics may prove especially valuable when assessing potential benefits, for instance, in the context of correction-based inference strategies such as speculative decoding~\cite{speculative}.
App.~\ref{sec:compl_gt} further discusses the use of ground-truth versus model generated probes.
We now empirically demonstrate the improvements of well-known compression methods using our metrics.


\section{Token Metrics Improve Model Compression}
\label{sec:compression}
We will demonstrate in the following how the proposed metrics provide novel insights into the efficiency of the architecture of LLMs and establish benchmarks for model compression. Throughout all experiments, we outperform standard PPL as a ranking metric.

More precisely, we apply componentwise probing to sparsification to determine individual sparsity rates. Interestingly, the model tends to completely remove components of the attention mechanism in certain layers. In total, 40 out of 160 attention components are sparsed beyond 90\% and 15 removed completely. For quantization, on the other hand, we show how component selection significantly influences the overall number of model outliers. For the first time, almost 10\% of the model components can be converted to 4-bit integer representations without significant model degradation.



\subsection{Experimental Protocol} 
\label{sec:exp}
Let us start by clarifying the experimental setup.

\paragraphown{Test environment.}
All experiments were performed on the public Llama2-7B and 13B models~\cite{llama2}. 
Note, however, that we observed similar behavior amongst other decoder transformer models. It remains, in particular, with upscaled sizes and smaller variations on the architecture or the training procedure. 

For all experiments, we follow best practices for compression evaluations \cite{sun2023wanda} and randomly sample data from the C4 dataset~\cite{c4dataset} for training iterations. The final model evaluation is performed comparing the loss on the Wikitext2 dataset~\cite{merity2016pointer} and the standard NLP benchmarks~\cite{eval-harness}.

\paragraphown{Metrics.} 
We apply our proposed metrics for performance evaluation, as well as selection criteria. 
We employ FDT, SDT, DPPL and PPL as metrics to assess the overall model divergence. With regard to model compression, we demonstrate that both PPL and our variant DPPL typically struggle to measure minor changes adequately (\textit{cf.}~Sec.~\ref{subsec:tokenvsperplexity}, \ref{sec:sparse} and Fig.~\ref{fig:sparse_random}). However, FDT is particularly suited to characterize errors for subtle model changes. Consequently, we apply FDT for model compression. In the following paragraph, we describe the parameters selected for using FDT in more detail.

\paragraphown{Divergent Token Parameters.}
We empirically selected hyperparameters as follows.
Through preliminary sparsification experiments, we observed that the most variance is present in the 75\%-quantile of FDT, as defined in Eq.~\ref{e:fdt} (\textit{cf.} Fig.~\ref{fig:prune_component}).
We denote this value by \emph{FDT$_{75}$}. 
In the following, it is our compare-to value. 

\begin{figure}[t]
\centering
\includegraphics[width=0.75\linewidth,height=100px]{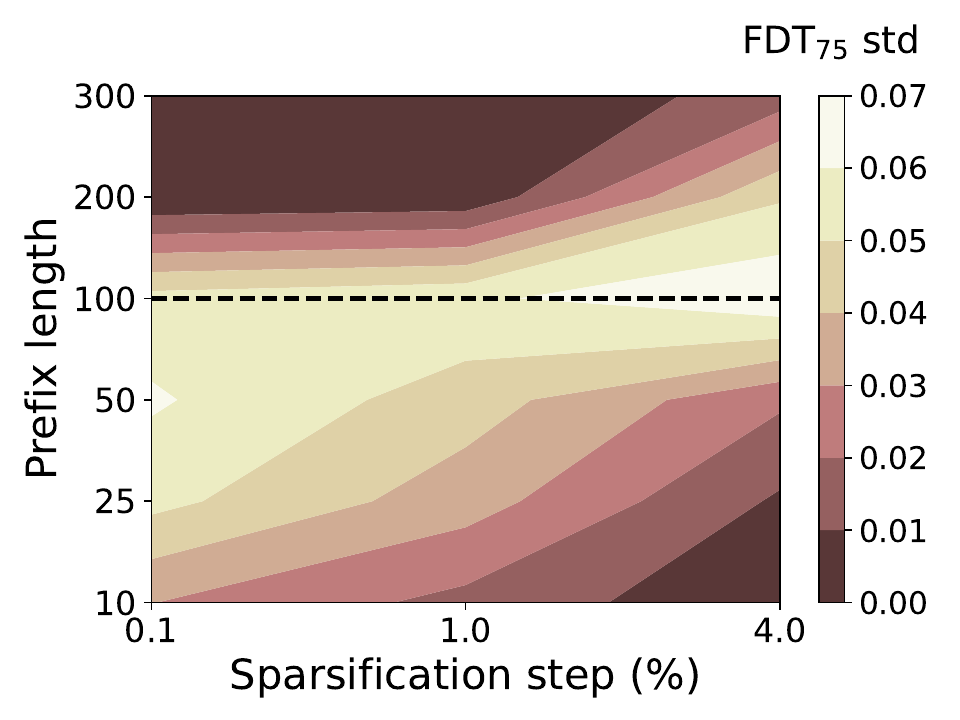}
\caption{Hyperparameter selection of FDT. 
Visualized is the standard deviation (std) in FDT$_{75}$ over all components when varying prefix length (y-axis) and applying different choices for sparsity-step increases (x-axis), \textit{cf.}~Sec.~\ref{sec:exp} and \ref{sec:sparse}.}
\label{fig:params}
\end{figure}

Next, we sweep over the given \emph{context prefix length} $n$ of FDT and sparsification steps as depicted in
Fig.~\ref{fig:params} on the y- and x-axis, respectively.
The heatmap shows the overall standard deviation of FDT$_{75}$ on 5k probes.
For simplicity, we fixed the prefix length to $100$ tokens, as it is most discriminative on average.

We observed that most of the sparsification steps introduce an error in FDT$_{75}$ within the range of $100$ \emph{completion tokens}. Therefore, we selected $N=100$.
Finally, to determine the \emph{number of probes} $|\mathcal{D}|$, we compared the mean deviation with a baseline of $5000$ probes. 
As the deviation in the value of $\mathcal{M}_{F_{75}}$ of $1000$ to $5000$ probes only differs on average by a value $4$, we selected $|\mathcal{D}|=1000$ (\textit{cf.} Fig.~\ref{fig:boxplot_1k5k}).


\paragraphown{Pruning of LLMs.}
\begin{algorithm}[h]
    \caption{Iteration of pruning algorithm}\label{alg:pruning}
\begin{algorithmic}
\State \textbf{input:} $F, step$ \Comment{current model, target sparsity}\vspace{4px}
\State $\textrm{fdt\_sparse\_map}\gets \{\}$
\For{$c_i \in \textrm{Components}$}
\State $\textrm{fdt\_sparse\_map}[c_i] \gets [100,$ 
\State\phantom{aabd}$\mathcal{M}_{\textrm{FDT}_{75}}(F, F^{c_i+step/2}), $  \State\phantom{aabd}$\mathcal{M}_{\textrm{FDT}_{75}}(F, F^{c_i+step+step/2}), 0]$\vspace{3px}
\\
\Comment{FDT values measured for added sparsities of  \\
\phantom{afds}$0$, $step/2$, $step+step/2$, $100\%$ to \\
\phantom{afds}component $c_i$ on model $F$. The maximal \\
\phantom{afds}FDT value measured is $100$.}
\EndFor\vspace{4px}
\State $f \gets 100$ \Comment{iteratively decrease from maximum} 
\State $s \gets 0$ \Comment{track current added sparsity}
\State $\textrm{comp\_sparse\_map} \gets \{\}$\vspace{4px}
\While{$s \leq step$ and $f \geq 0$}
\For{$c_i \in \textrm{Components}$}
\State $\textrm{comp\_sparse\_map}[c_i] \gets$ 
\State \phantom{afds}$lin\_interpol(\textrm{fdt\_sparse\_map}[c_i], f)$
\EndFor
\State $s \gets weighted\_mean(\textrm{comp\_sparse\_map})$
\State $f \gets f -1$
\EndWhile \vspace{4px}
\State $F' \gets F$ pruned by \textrm{comp\_sparse\_map} \vspace{4px}
\State \textbf{output:} $F'$
\end{algorithmic}
\end{algorithm}
In Sec.~\ref{sec:sparse}, we will show that FDT improves sparsification procedures to achieve high compression rates on LLama2-13B. 
To this end, we iterate small unstructured sparsification with continued training steps for the model to attune to the remaining weights and recover performance.
Specifically, we apply eight iterations to increase the average model sparsity by $20,15,10,10,5,5,5,$ and $5$ percent, resulting in a final model with 25\% total parameters remaining.

We run this experiment in two configurations, uniform and FDT-selective.
Uniform sparsification applies the target increase of the current round uniformly to each component, pruning their lowest weights.
For FDT, we determine individual component sparsification values to evenly distribute the induced error as depicted in Algorithm~\ref{alg:pruning}: 

Based on the previous sparsed model $F$ and for the current target increase $step$, we probe each component $c_i$ separately with an additional $step\pm step/2$ percent of the lowest weights pruned, denoted by $F^{c_i+s}$, to determine its $\mathcal{M}_{FDT_{75}}$ value. We further add the constant extrema, that is, step sparsity $0$ and $100\%$ with values of $100$ and $0$.
Given these four data points, we segment-wise interpolate linearly to achieve the highest value of $\mathcal{M}_{FDT_{75}}$ possible across all components, but on average yielding the target sparsity. 
Specifically, we find the set of component-sparsities $\{s_i\}$ that optimize for
\begin{align*}
    \textstyle\argmax_{\{s_i\}} \min_{i} \mathcal{M}_{\textrm{FDT}_{75}}(F, F^{c_i+s_i}),
\end{align*}
such that $\sum_i \tilde{s}_i = step$ using $\tilde{s}_i$ to represent the normalized sparsity of $s_i$ relative to the individual parameters of component $c_i$.

We further follow the findings of AC/DC~\cite{peste2021ac} and alternate compressed and decompressed iterations as follows: Each round we train a total of $500$ steps, of which the first 450 are with the sparsification mask applied and the following $50$  without any masks. We found that this alternation produces smaller spikes in training loss after sparsification steps. 

This yields a total of $4000$ training steps.
During training, we apply a weight decay of $0.01$, batch size $256$, and sequence length $2048$. 
Note that throughout this series of experiments, we only apply pure magnitude pruning per iteration. 
Other pruning methods, such as wanda~\cite{sun2023wanda}, can also be straightforwardly augmented by this probing strategy. 

\begin{figure}[t!]
\centering
\begin{subfigure}{\linewidth}
\includegraphics[width=\linewidth]{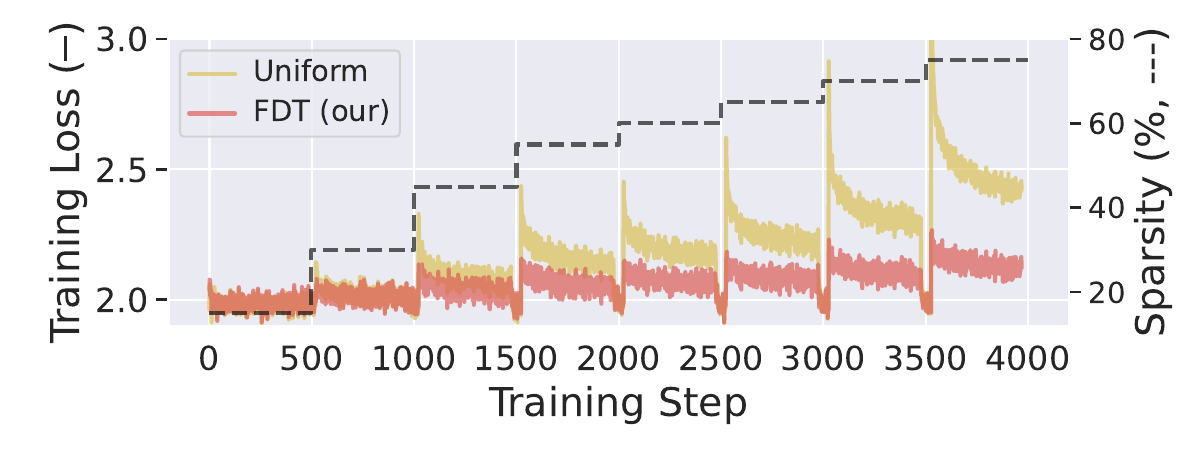}
\caption{Comparison of uniform and componentwise pruning using FDT as a metric for comparison. }
\label{fig:loss_sparse}
\end{subfigure}
\begin{subfigure}{\linewidth}
    \centering
\includegraphics[width=\linewidth]{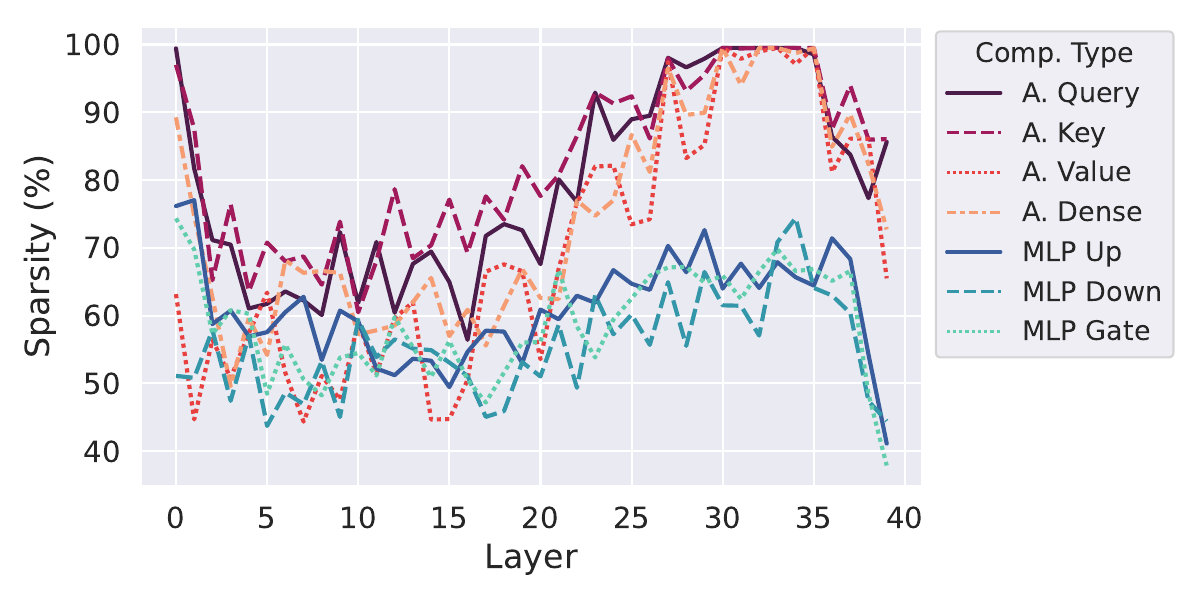}
    \caption{Converged component config with 75\% average sparsity. Layers (x-axis), Component-sparsity (y-axis).}
    \label{fig:sparse_componentwise}
\end{subfigure}
\caption{
Depiction of the proposed sparsification process that converged to a 75\% sparse Llama-2-13B.
\textbf{a)} Model training performance throughout all rounds. Our FDT-based sparsification clearly outperforms uniform magnitude pruning.
\textbf{b)} Converged sparsity values per component.
One quarter of attention components are pruned beyond 90\% sparsity.
Significant outliers appear in first and last layers.
}
\label{fig:sparse_80}
\end{figure}
\paragraphown{Quantization of LLMs.}
For model quantization, we compare the performance of the proposed metrics on the task of sorting the model's components by their lowest introduced error.
To this end, we build a search tree to find the best model configuration as follows: 
We build a parallel depth-first search tree with a branching width of $10$, which relates to a beam search with beam width $10$. In particular, at each level of the tree, we simultaneously explore all possible successor configurations for the currently top-$10$ performing nodes, with one more component naively quantized using AbsMax. From this newly identified set of nodes, we again select the best performing $10$ nodes for the next iteration. Starting with the unquantized base model Llama2-7B, each node contains exactly the number of quantized components respective to its depth, while the final node is a fully AbsMax quantized model. We further apply deduplication to prevent redundant computations.


\begin{figure}[t]
    \begin{subfigure}{\linewidth}
\begin{center}
\includegraphics[width=0.9\linewidth,height=100px]{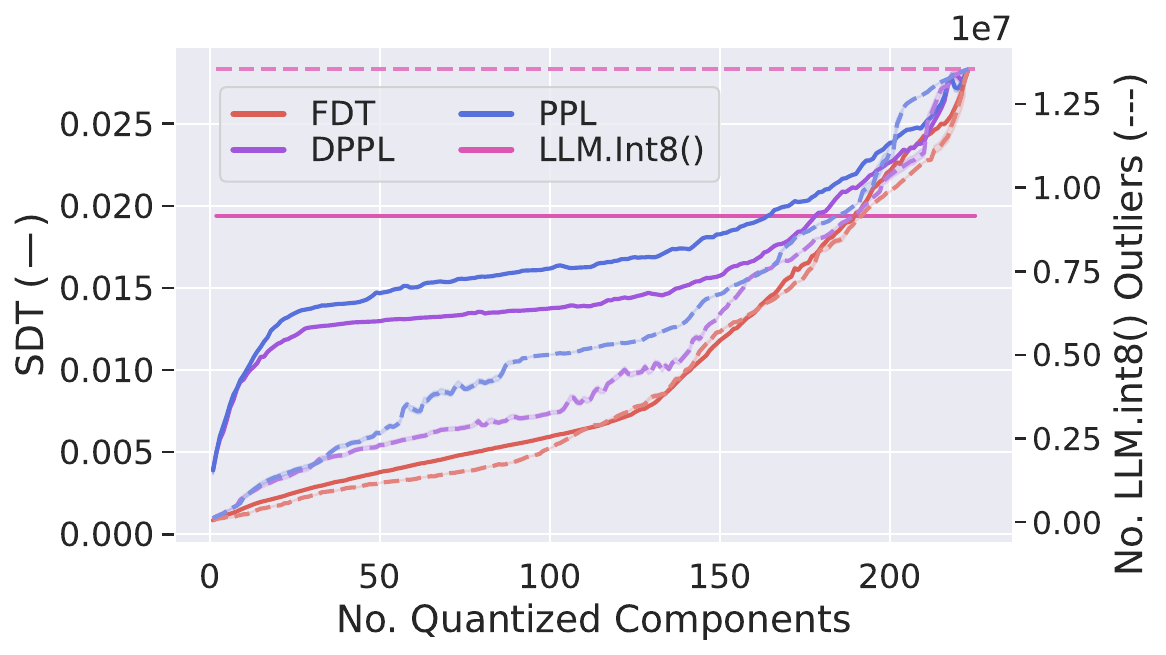}
        \caption{Performance of FDT, DPPL and PPL.}
    \end{center}
    \end{subfigure}
    \begin{subfigure}{\linewidth}
        \begin{subfigure}{\linewidth}
\begin{center}
            \includegraphics[width=0.9\linewidth]{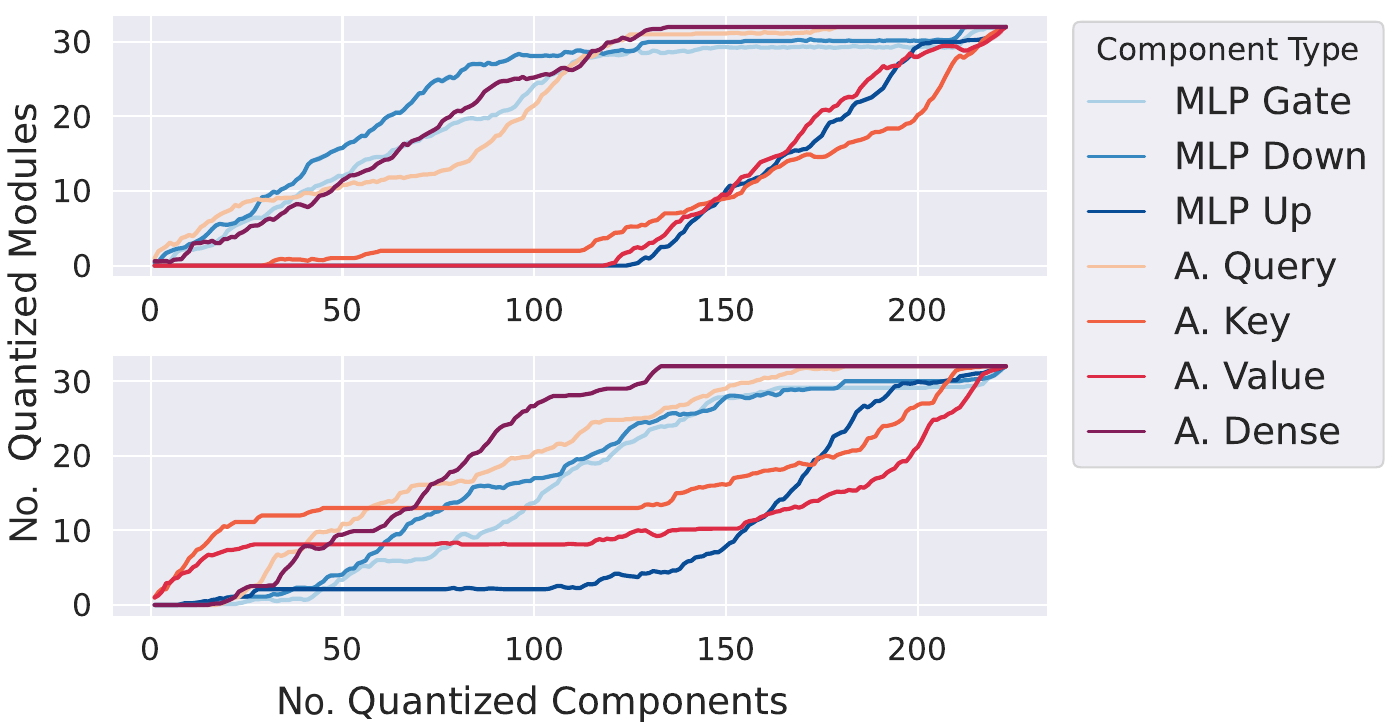}
            \caption{Selected Components FDT (top) vs PPL (bottom).}
            \label{fig:search_tree_split}
    \end{center}
        \end{subfigure}
    \end{subfigure}
    \caption{ Evaluation of the Tree Search as described in text.
    \textbf{a)} Comparison of Tree Search based componentwise quantization. 
     Different numbers of components (x-axis) lead to different token divergence scores (y-axis, normalized to $[0,1]$), and in particular correlates early on to introduced outliers (second y-axis).
    Throughout the entire search, FDT is able to rank components by their potential errors and, coincidentally,  outliers.
    \textbf{b)} Selected components at respective depth. A.Key and A.Value induce most error.}
    \label{fig:search_tree}
\end{figure}
\subsection{Sparsification reveals: \\ \hspace{22px} Attention is not all you need!}
\label{sec:sparse}

We applied stepwise uniform magnitude pruning, and our balanced componentwise pruning using FDT, to achieve 75\% model sparsity.
A summary of the results is shown in Fig.~\ref{fig:sparse_80}.



\paragraphown{Attention almost erased.}
Fig.~\ref{fig:sparse_componentwise} visualizes the converged sparsity values when applying our balanced pruning using FDT.
%
%
Notably, the model favors pruning attention over MLP. In total 40 of 160 attention components are sparsed beyond 90\% and 15 even completely removed. 
In general, the second half of the model appears to be more prunable than the first half. The value matrices are overall the least pruned of the attention matrices.
Finally, significant outliers appear in the first and last layers.
This finding indicates that attention is not utilized efficiently throughout model inference.
In fact, only layers 3 to 20 and layer 40 appear to be of significant relevance for the model's final prediction. This observation might be attributed to an evolving shift in distributions and, with that, the concepts processed in embeddings.

Notably, in the first layer Attention Value and MLP Down remain significantly dense, while all others are comparably sparse. This observation indicates an incomplete shift of token embeddings.


\paragraphown{General Observations.}
\begin{table}[t]
    \centering
    \small
    \setlength{\tabcolsep}{5.5pt}
    \begin{tabular}{cl|r|r|r}
        \multicolumn{5}{c}{Sparsification}\\
        &Model & FDT $\uparrow$ & PPL $\downarrow$ & NLP $\uparrow$ \\ \hline
        &Llama2-13B & - & \phantom{}4.884 &  53.59 \\\hdashline
        &$\sim$ 60\% sparse (unif.)& \phantom{00}\phantom{}4.7 & \phantom{0}\phantom{}9.244 &  46.32 \\
        &$\sim$ 60\% sparse (our)& \phantom{00}{}\textbf{7.9} & {}\textbf{6.242} & \textbf{48.89}  \\\hdashline
        &$\sim$ 75\% sparse (unif.)& \phantom{00}\phantom{$\bullet$}3.5 & \phantom{$\bullet$}13.512 &   41.67
        \\
        &$\sim$ 75\% sparse (our)& \phantom{00}{}\textbf{5.5} & {}\textbf{8.101} & \textbf{46.32}   \\\hdashline
        &$\sim$ 80\% sparse (our)& \phantom{00}{}5.2 & {}9.531 & 45.66\\ 
        \multicolumn{5}{c}{} \\ 
        \multicolumn{5}{c}{Quantization}\\
        &Model & FDT $\uparrow$ & PPL $\downarrow$ & NLP $\uparrow$ \\ \hline
        &Llama2-7B & - & \phantom{}5.472 & 50.79\\ \hdashline
        \parbox[t]{1mm}{\multirow{4}{*}{\rotatebox[origin=c]{90}{int8}}} & LLM.int8()$_{all}$ & \phantom{0}\phantom{}36.1 & \phantom{}5.505 & \textbf{50.81}\\ 
        &AbsMax PPL$_{150}$ & \phantom{0}\phantom{$\bullet$}46.3&  \phantom{}5.500& 50.72\\ 
        &AbsMax DPPL$_{150}$ & \phantom{0}\phantom{}54.1&  \phantom{}5.490& 50.75\\ 
        &AbsMax FDT$_{150}$  (our)& \phantom{0}\textbf{71.7} & \textbf{5.489} & 50.75 \\ \hdashline
        \parbox[t]{1mm}{\multirow{4}{*}{\rotatebox[origin=c]{90}{int4}}}& GPTQ$_{all}$ & \phantom{}\phantom{0}11.1 & \phantom{}5.665 & 48.34 \\
        &GPTQ PPL$_{16}$  & \phantom{} 45.0 & \phantom{}5.511& 49.91 \\
        &GPTQ DPPL$_{16}$  & \phantom{}\phantom{0}137.0 & \phantom{}5.476& 50.02 \\
        &GPTQ FDT$_{16}$  (our)&\textbf{205.0} &\textbf{5.475} & \textbf{50.13} \\
    \end{tabular}
    \caption{Evaluations of the final Compressed Models. Even when evaluating these, aggregations of standard NLP benchmarks (\textit{cf.} Fig.~\ref{fig:quant_nlp},\ref{fig:sparse_nlp}) do not reflect the actual degradation of the model, as observed in AbsMax quantization. FDT and PPL are evaluated on Wikitext2 samples. Throughout all experiments, our FDT probed compression outperforms standard variations. 
    Subscript refers to best found $k$ quantized components. Bold denotes the best values.}
\label{tab:comp_model_benchmarks}
\end{table}
\begin{figure*}[t!]
\centering
\begin{subfigure}{.6\linewidth}
\includegraphics[width=\linewidth,height=80px]{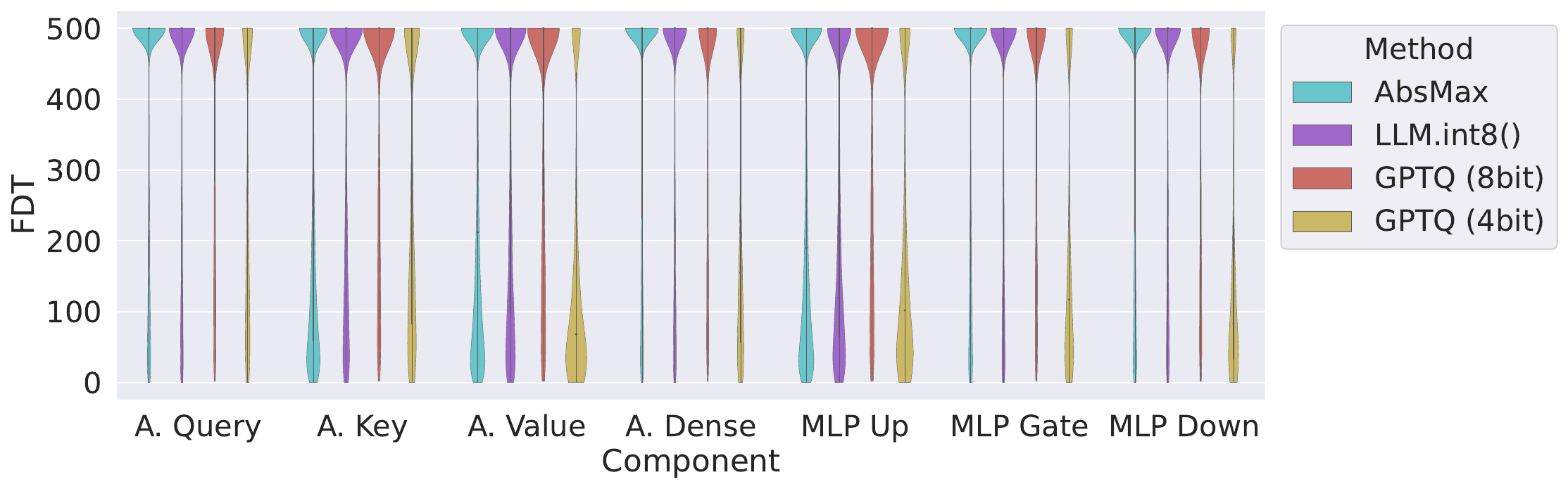}
\caption{Quantization methods evaluated on Components.  }
\label{fig:fdt_dpplvsfdt}\label{fig:comp_quant_violin}
\end{subfigure}
\quad
\begin{subfigure}{.36\linewidth}
\includegraphics[width=\linewidth,height=80px,clip,trim={0 1.4cm 0 0}]{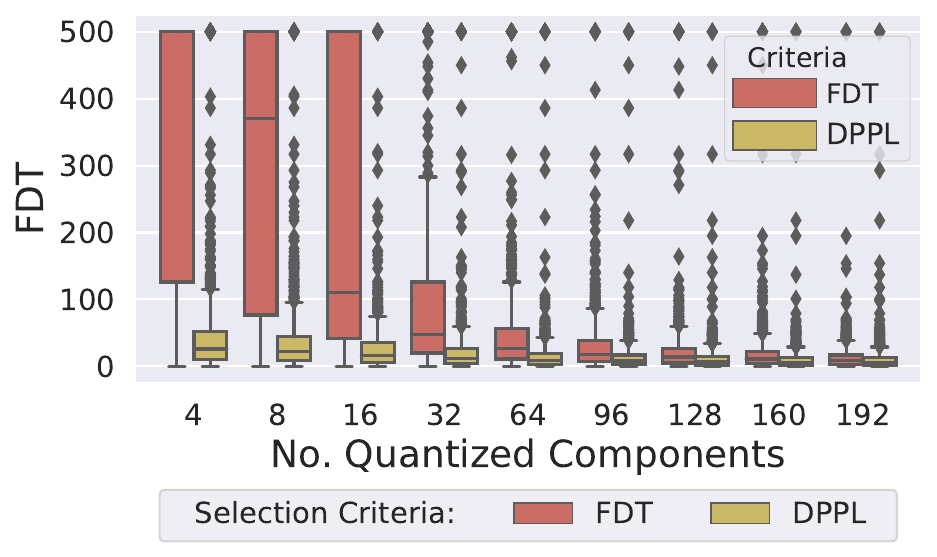}
\caption{Selected GPTQ (4bit) components.}
\label{fig:fdt_methodwise}\label{fig:select_gptq4}
\end{subfigure}
\caption{Evaluation of FDT performance. 
\textbf{a)} evaluates components separately on all quantization methods. Clear outliers in performance are A. Value and MLP Up. GPTQ (8bit) is able to evenly amortize the induced error.
\textbf{b)} Selecting top-k components of GPTQ (4bit). 
FDT is suited to rank components one-shot.}
\end{figure*}
%
%
%
As shown in Fig.~\ref{fig:loss_sparse}, FDT based balanced pruning significantly lowers the introduced error between sparsification rounds.
Uniform pruning, on the other hand, substantially diverged, and in particular does not regain performance with the given amount of compute. 
Generally speaking, what is lost can hardly be recovered. 


The standard evaluation of FDT and PPL on Wikitext2, is found in Tab.~\ref{tab:comp_model_benchmarks}. The  75\% compressed 13B model,  with several components pruned away,  scored PPL 8.1, compared to PPL 4.8 of the base model. Note that no other model sparsed beyond 70\% has yet been reported, in particular, achieving single-digit PPL. Uniform pruning achieved 13.5. Further note, that we almost doubled the mean FDT value when compared to uniform pruning.
However, as the generally low FDT value suggests, it still diverged substantially from the base model.






\paragraphown{FDT is more discriminative.}
In practice, FDT is better able to discriminate subtle changes than PPL.
We demonstrate this with the following experiment: On each component of the model, we prune 0.1\% of the weights either randomly or from the lowest weights.
The resulting model is probed for $1000$ trials with all the discussed metrics used to distinguish the cases.
The results in Fig.~\ref{fig:sparse_random} clearly indicate that FDT is capable of distinguishing the cases, while they remain indifferent for PPL-based comparison.
We therefore omit using PPL as a metric to determine step-sizes for the described sparsification experiment.
%

\subsection{Quantization: Outliers can be prevented}
\label{sec:quant_tree}

Finally, we demonstrate the impact of selecting the right components for quantization.
We compare the proposed metrics PPL, DPPL, and FDT as ranking criteria to showcase their discrimation capabilities.


%
%
%
\paragraphown{Quantization without outlier-handling.}
%
%
%
%
%
%
%
%
Fig.~\ref{fig:search_tree} shows the average performance of the top 10 nodes occuring in the respective search tree depth (x-axis). 
FDT continuously outperforms the other metrics on the SDT value (y-axis).
Notably, this is on par with the total number of outliers occurring for the respective configurations (second y-axis).
Certain components appear to significantly influence the decline observed in both measures. Although DPPL enhances some aspects of performance, neither variant of PPL effectively distinguishes these components and tends to select those prematurely.

With FDT, we can cast 80\%, \textit{i.e.}~150, of the model's components directly into int8 using only naive AbsMax---and without further outlier handling---still outperforming the full LLM.int8() conversion in model performance. 
Selecting those 150 components with DPPL and FDT leads to close perplexity scores $5.490$ and $5.489$ on Wikitext2, \textit{cf.} Tab.~\ref{tab:comp_model_benchmarks}.
However, the resulting mean FDT improves by almost 50\% when also selecting the components by this metric. The larger generation of the same sequences suggests a model closer to the original when choosing FDT as a selection criterion.
%
%
Fig.~\ref{fig:search_tree}b) shows the selected components at each depth in relation to a).
 Most outliers occur when the Attention Key is selected early on. Notably, we observed in Sec.~\ref{sec:sparse}, that this is one of the matrices most suitable to sparsify. 

%
%
\paragraphown{16 components in 4-bit.}
Figure~\ref{fig:comp_quant_violin}) presents a comprehensive assessment of the quantization techniques discussed. 
First, it is noticeable that the LLM.int8() method slightly improves the lower quantile scores of FDT in comparison to AbsMax. 
However, GPTQ (8bit) demonstrates superior performance, outshining both plain AbsMax and LLM.int8(). This method achieves a more balanced error distribution across all components (\textit{cf.}~App.~Fig.~\ref{fig:componentwise_plots}). 
On the contrary, GPTQ (4-bit) shows noticeable deviations in the generation process, with only a limited number of components achieving FDT scores greater than 400. Despite this, the discriminative power of FDT enabled us to identify and merge the top 16 components that minimally compromised the integrity of the model, as illustrated in Fig. \ref{fig:select_gptq4}).

\section{Conclusion}
\label{sec:conclusion}
We introduced the Divergent Token Metrics (DTMs), a tailored approach to evaluate the performance differences of compressed generative models.
In particular, DTMs respect the usually applied greedy sampling procedure to generate predictions.
We proved that DTMs achieve appropriate metric bounds and are not affected from catastrophic artefacts that perplexity-based metrics encounter. 
We constructed algorithms for iterative sparsification and quantization processes based on DTM probing of individual components.
Using DTMs, we achieved an outperforming  75\% sparse version of Llama2-13B and successfully converted 80\% of the LLama2-7B  components naively to int8.

\section*{Limitations}
With the proposed DTMs, compression processes can be tailored to use cases---and we can measure their performance degeneration.
We hinted with the sparsification experiments, that MLP and Attention can be ascribed varying levels of significance throughout the layers. These studies should be further extended to various model architectures such as BERT or MoE.
Moreover, variations of specific datasets to probe or finetune on could lead to interesting observations. For example, multilingual, multitasking and instruction tuned MoE models could be pruned multiple times using FDT to investigate how the information actually is stored and assembled. Note that this work solely relied on English-tailored pruning for a mostly English-trained model.
Furthermore, augmenting our metrics to probe for (and investigate) specific aspects such as safety alignments, and preserving these elements throughout model compression would be a significant area of study.

As a pruning strategy, we achieved outperforming results using only naive magnitude pruning. 
DTMs should be directly applicable to other masking strategies, such as Wanda~\cite{sun2023wanda}, which may further improve results.
Finally, the generalizability of the metrics to other sampling strategies should  be investigated.



\section*{Acknowledgments}

We thank the three anonymous referees for their constructive comments and suggestions, which helped us to improve this paper considerably.

This work has been partially funded by the Deutsche Forschungsgemeinschaft (DFG, German Research Foundation) as part of BERD@NFDI - grant number 460037581.

We gratefully acknowledge support by the German Center for Artificial Intelligence (DFKI) project “SAINT”,
the Hessian Ministry of Higher Education, the Research and the Arts (HMWK) cluster projects
“The Adaptive Mind” and “The Third Wave of AI”, and the ICT-48 Network of AI Research Excellence Center “TAILOR” (EU Horizon 2020, GA No 952215). 

We thank Graphcore, Jamie Packer, Manuel Brack and Samuel Weinbach for the fruitful discussions and the invaluable feedback throughout this work.


\bibliography{references}

\clearpage

\appendix

\section*{Appendix}

\section{Proof of Propositions}
\label{sec:proofs}
\begin{proof}[Proof of Proposition~\ref{prop:ppl_discontinuity}]
    There are many ways to construct sequences that satisfy the desired relation. One is as follows: Let $l \in \R^{N \times |\mathcal{V}|}$ be any logit sequence with no re-occurring values. Denote by $m_k(l)_i$ the top-$k$ value  at position $i$, and by $a_k(l)_i$ the top-$k$ vocab index at position $i$, respectively. Now we pick any such sequence with the additional property that $\max_i m_1(l)_i - m_2(l)_i < \delta$ for some small $\delta$. Define the sequence $l'$ by
    \begin{align*}
        l'_{ia_2(l)_i} &= l_{ia_2(l)_i} + \delta,
    \end{align*}
    and $l'_{ij} = l_{ij}$ for all remaining indices. Then we have $a_1(l)_i \neq a_1(l')_i$ for all $i$ and hence $M_{\mathrm{SDT}}(l,l',y_{:1}, N) = N$. On the other hand we have $||l-l'||_\infty \leq \delta$. Since $\operatorname{PPL}(y,l,1)$ is a continuous function in $l$, we have $|\operatorname{PPL}(y,l,1) - \operatorname{PPL}(y,l',1)| < \varepsilon$ for any $\varepsilon$ and small enough $\delta$.
\end{proof}

\begin{proof}[Proof of Proposition~\ref{prop:upper_bound}]
 Let  $z = \mathcal{G}(l,y_{:n},N)$ and $p_i = (\mathrm{softmax} \ l')_{iz_{i+1}}$. Applying the definitions and elementary operations, we have
\begin{align*}
    \sum_{i=n}^N - \log p_i = (N-n) \log M_{\mathrm{DPPL}}(l,l',y_{:n}, N).
\end{align*}
Let $A = \{i \geq n \colon p_i \leq 1/2 \}$. Then
\begin{align*}
    \sum_{i=n}^N - \log p_i &= \sum_{i \in A} - \log p_i + \sum_{i \in A^c} - \log p_i \\
    & \geq \sum_{i \in A} - \log p_i \geq |A| \log 2 .
\end{align*}
Here we first used that $\log p_i \leq 0$ and then the observation that indices contained in $A$ satisfy $-\log p_i \geq \log 2$ by the defining property of $A$.
Finally, we argue that $ \operatorname{SDT}(z,l',n)\leq |A|$. Indeed, at any position $i$ where $\argmax_j l'_{ij} \neq z_{i+1}$ it must hold that $p_i \leq 1/2$, since any softmax-value larger than $1/2$ is automatically the maximum value of the distribution, and the softmax operation is monotone. Putting everything together we arrive at the desired inequality.
\end{proof}

\begin{figure}[t]
\centering
\includegraphics[width=.8\linewidth]{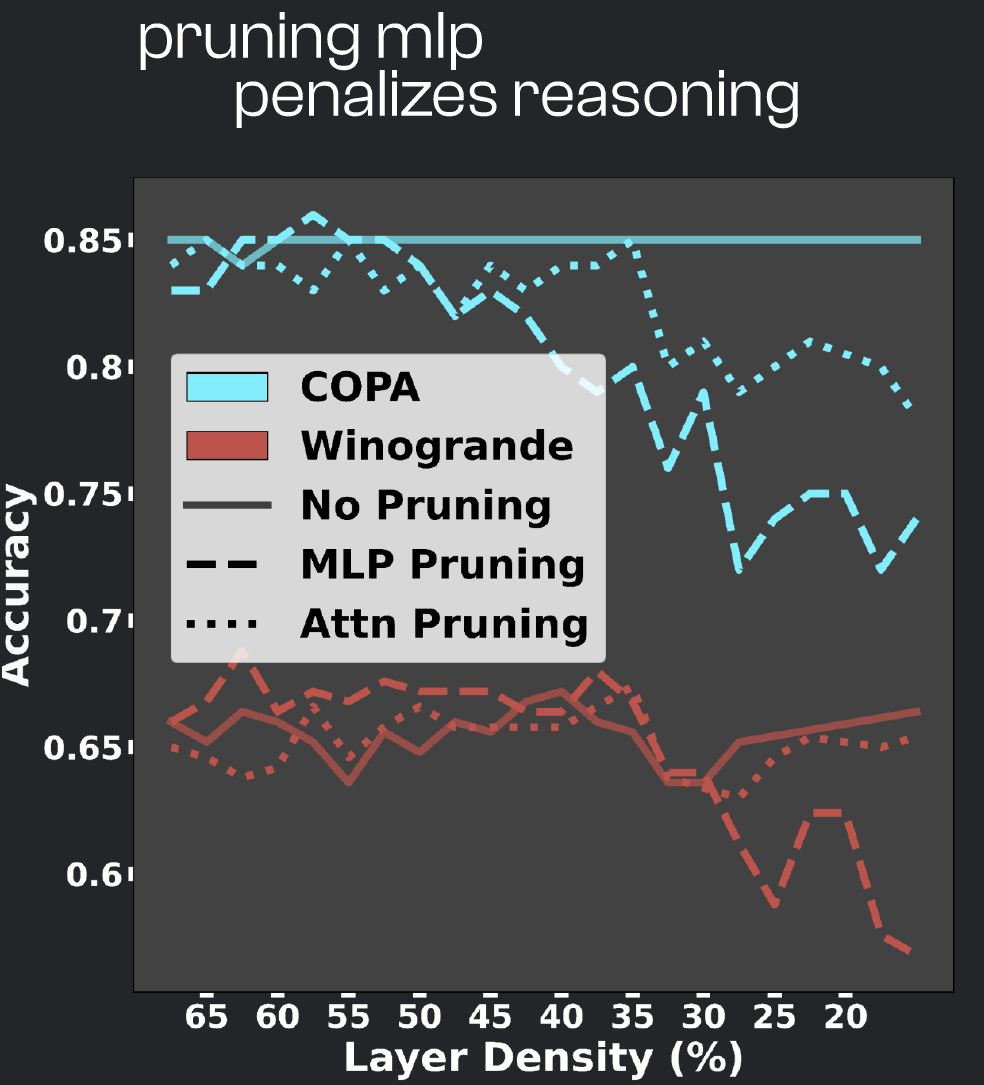}
\includegraphics[width=.8\linewidth]{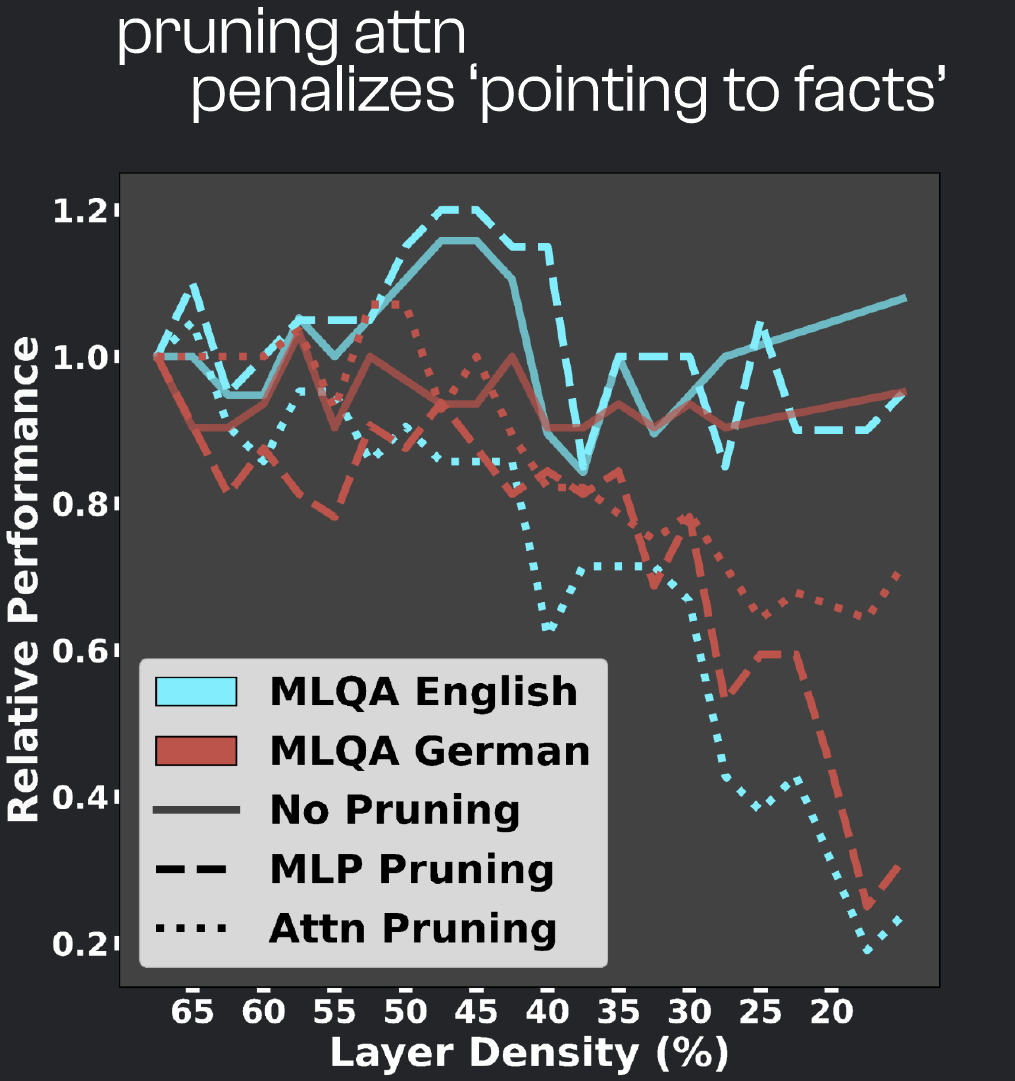}
\caption{Pruning MLP and Attn only indeed compromises remaining model capabilities.}
\label{fig:mlpvsattn_fig}
\end{figure}

\section{FDT compared to standard model evaluations}
\label{sec:quant_nlp_comparison}
\begin{figure}[t]
\includegraphics[width=\linewidth]{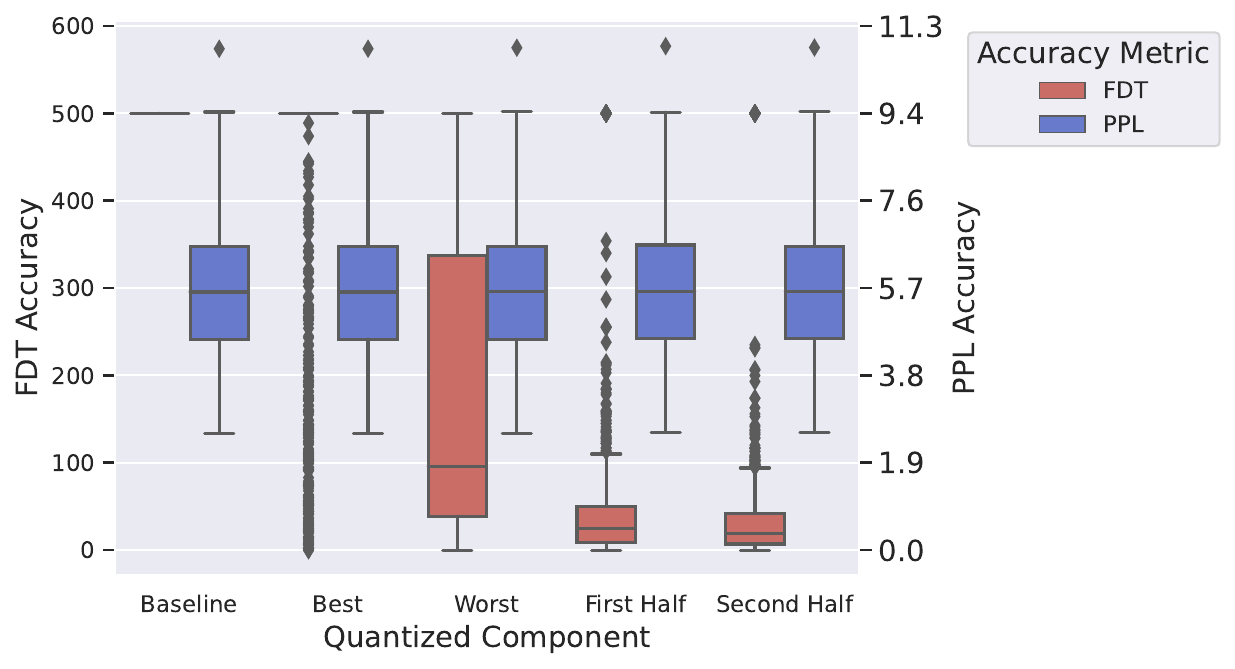}
\caption{Comparison of the discrimination capabilities of FDT and PPL for different configurations when applying LLM.int8() conversion on Llama2-7B. 
Best and Worst mark a single component being converted, with most and least mean influence.
First and Second half consecutively convert half of the model each.
While significant changes can be observed using FDT, all configurations appear indifferent for PPL.
}
    \label{fig:fdt_vs_nlp}
\end{figure}

Fig.~\ref{fig:fdt_vs_nlp} shows a comparison of standard benchmarks (middle) to FDT (right) and PPL (left) when quantizing parts of a model.
Often, standard evaluations fail to distinguish between compressed models. 
Sometimes they even depict better performance---which may be true, when regarding compression as a fine-tuning method and considering the short required token predictions.
FDT thoroughly gives discriminative statistics with resprect to the base model, on how much the compressed model equals the original. Note how the error seems to be upper bounded, which suggests that errors may average out throughout the model.
Mean zeroshot accuracy denotes the average on the standard NLP-eval harness.

\section{True positives can be predicted}
\begin{figure*}[h!]
\centering
\includegraphics[width=.6\linewidth,clip,trim={0 0 0 0}]{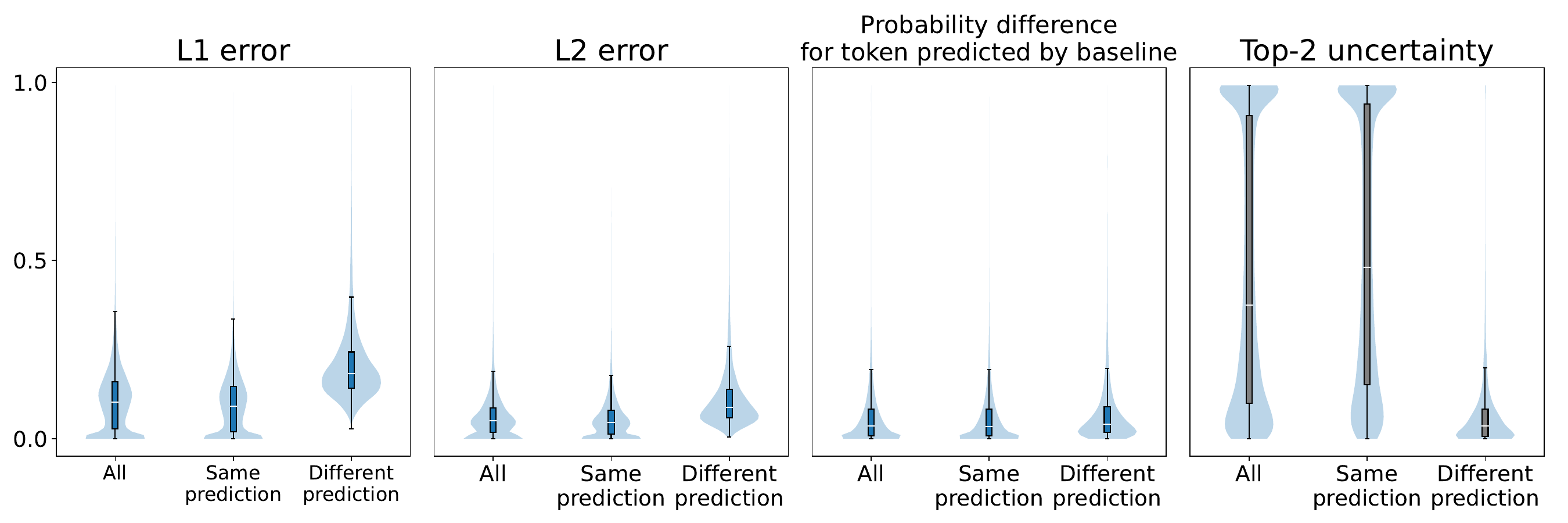}
\caption{Top-2 uncertainty is discriminative enough to give clear true-positives estimates on compressed models.}
    \label{fig:predicting_true_pos}
\end{figure*}

Fig.~\ref{fig:predicting_true_pos} shows several metrics applied to the token-distributions, in order to estimate on whether the compressed and original model predictions are equal. 
Notably, L1 and L2 errors on the entire distribution seem to somewhat capture the discriminative capabilities of false predictions. The probability scores themselves are only marginally usable.
Using top-2 uncertainty, i.e. the difference between the top-2 tokens as a measure, we obtain a reliable prediction of true positives.
True negatives however still remain with a significant overlap.

\section{MLP is for knowledge,\\\hspace{20px}Attention for relation}
\label{sec:mlpvsattn}

Finally, we observed that when pruning only attention, prompt-extraction capabilities degenerate severely. When only pruning MLP components, on the other hand, it influences mostly world knowledge QA benchmarks, \textit{cf.} Fig.~\ref{fig:mlpvsattn_fig}.

\section{Details on Search Tree, Sec.~\ref{sec:quant_tree}}
\begin{figure}[t]
\includegraphics[width=\linewidth]{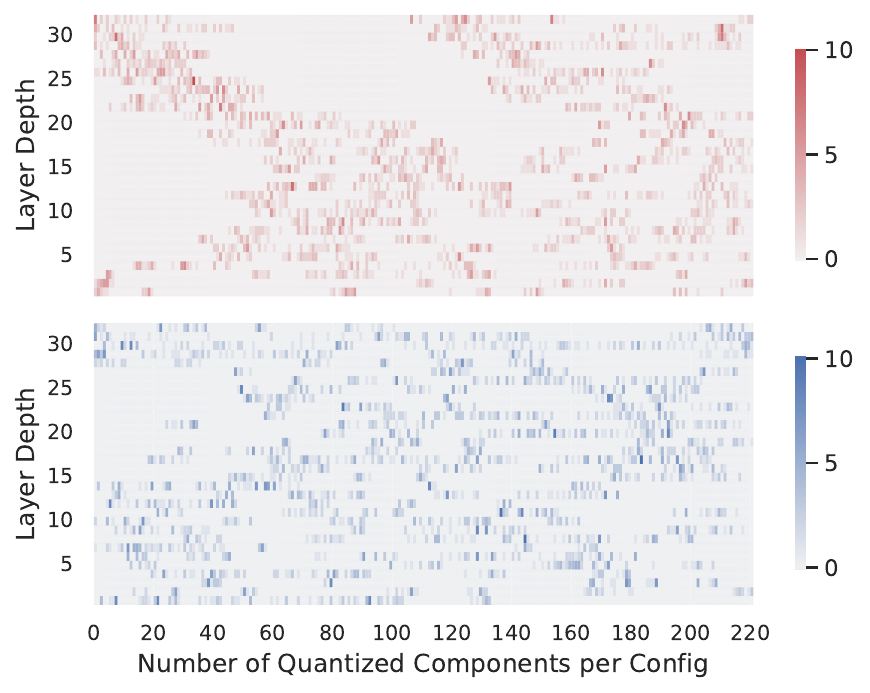}
\caption{Layers selected in each round of the search tree. Top, when applying FDT, bottom, when applying PPL as a ranking metric.}
\label{fig:search_tree_rounds}
\end{figure}

Fig.~\ref{fig:search_tree_rounds} shows the layers (y-axis) of which components are selected at each round (x-axis).
While there seems to be a pattern on when using FDT as a criteria (top), selection by PPL (bottom) looks more random.

Fig.~\ref{fig:comparison_search} shows the comparison of search tree as described to greedy search on a single evaluation of all components. Until 150 components, FDT proves more stable over the PPL variants as seen in Fig.~\ref{fig:comp_fdt}.


\begin{figure}[t]
\begin{subfigure}{\linewidth}
    \centering
    \includegraphics[width=\linewidth]{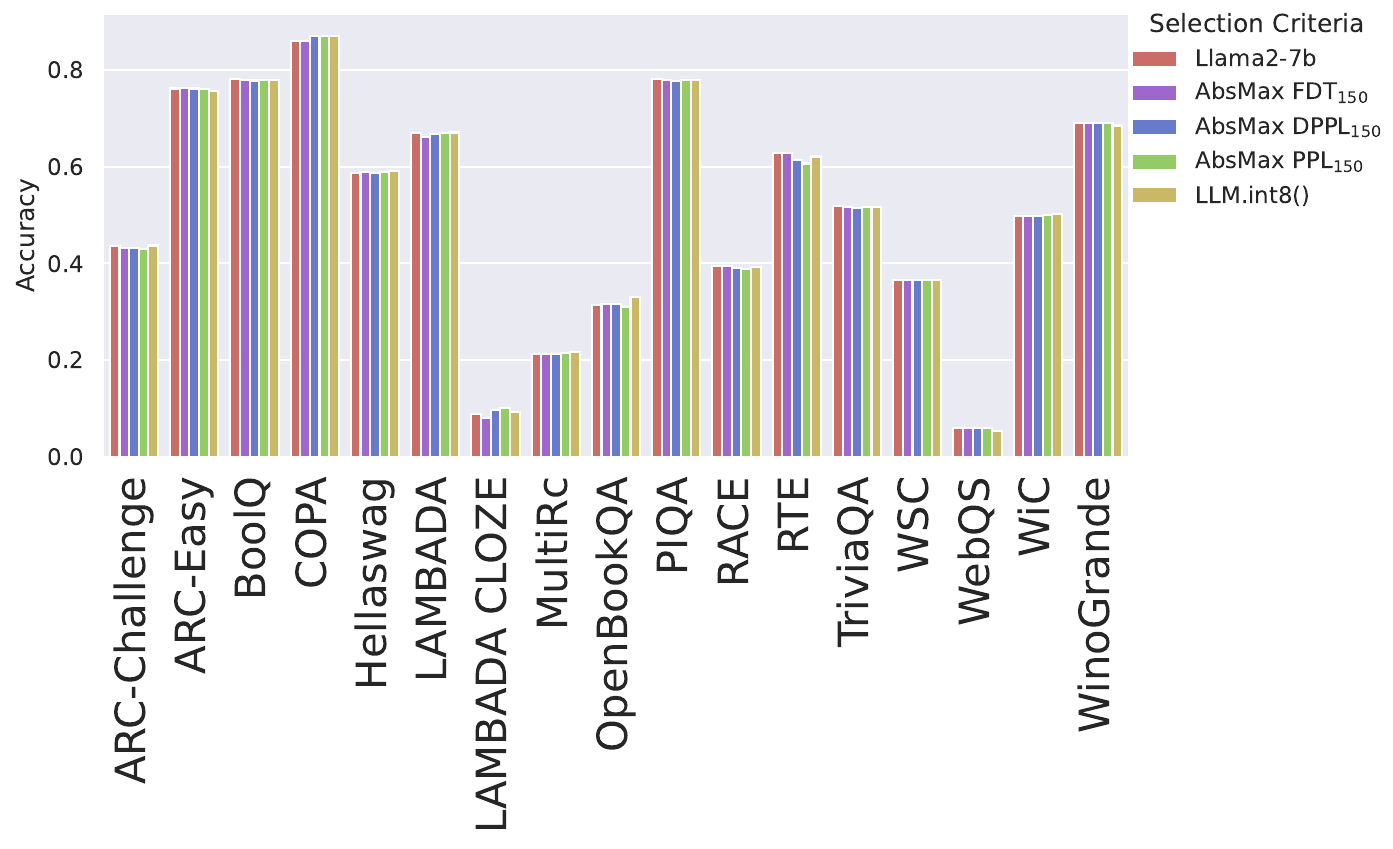}
    \caption{8-bit Quantization NLP benchmarks}
\end{subfigure}
\begin{subfigure}{\linewidth}
    \centering
    \includegraphics[width=\linewidth]{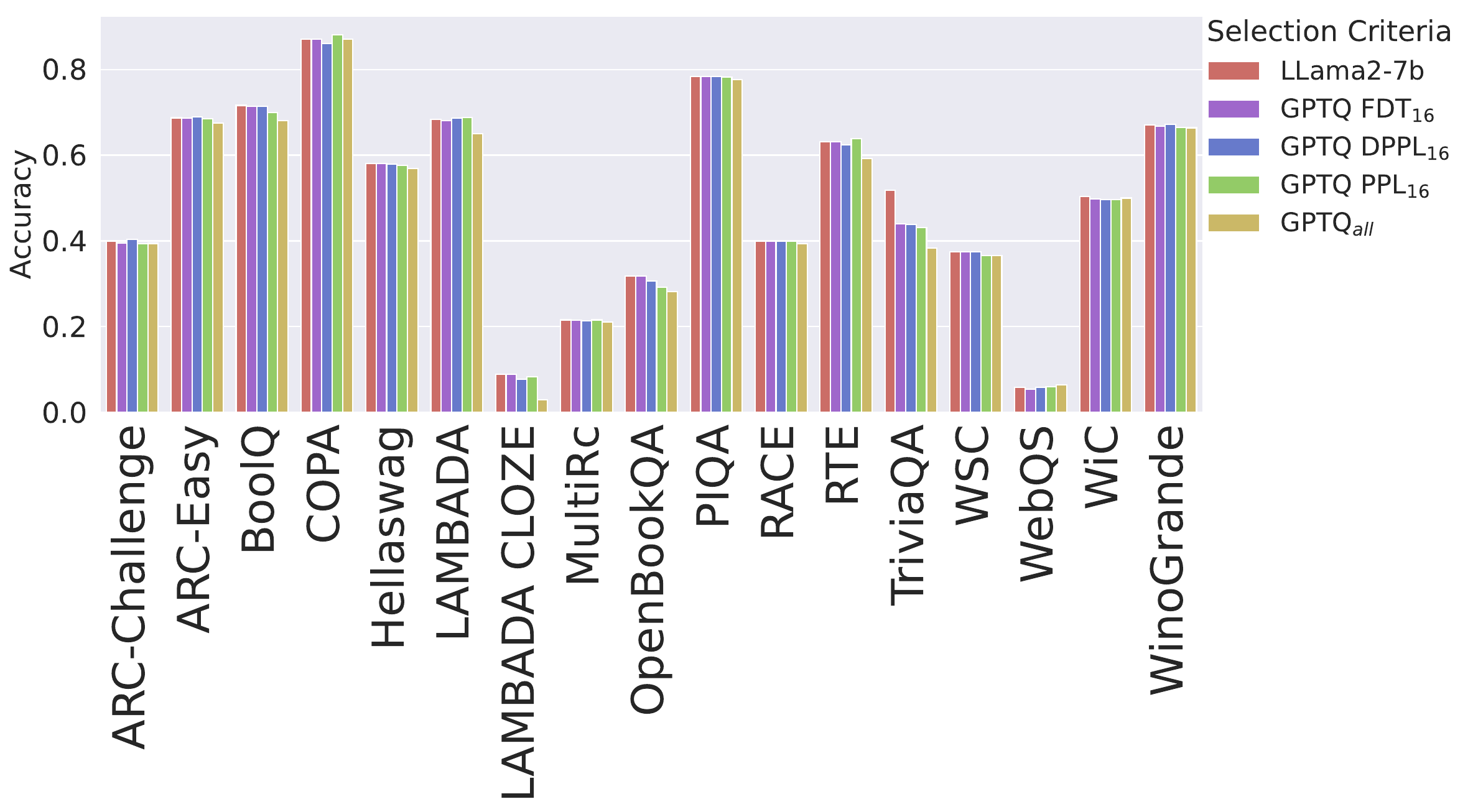}
    \caption{4-bit Quantization NLP benchmarks}
\end{subfigure}
\caption{Detailed view on aggregated values of Tab.~\ref{tab:comp_model_benchmarks} when selecting Llama2-7B components to quantize by metrics.}
    \label{fig:quant_nlp}
\end{figure}

\begin{figure*}[t]
\begin{subfigure}{0.49\linewidth}
    \centering
    \includegraphics[width=\linewidth]{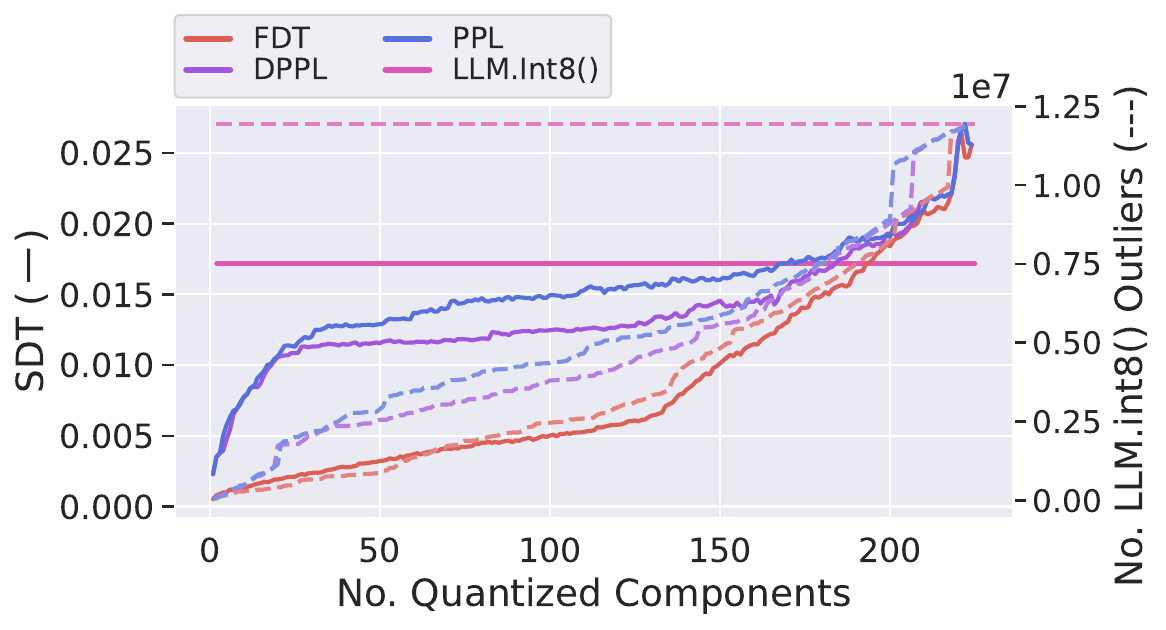}
    \caption{Context size of 10 tokens.}
\end{subfigure}
\begin{subfigure}{0.49\linewidth}
    \centering
    \includegraphics[width=\linewidth]{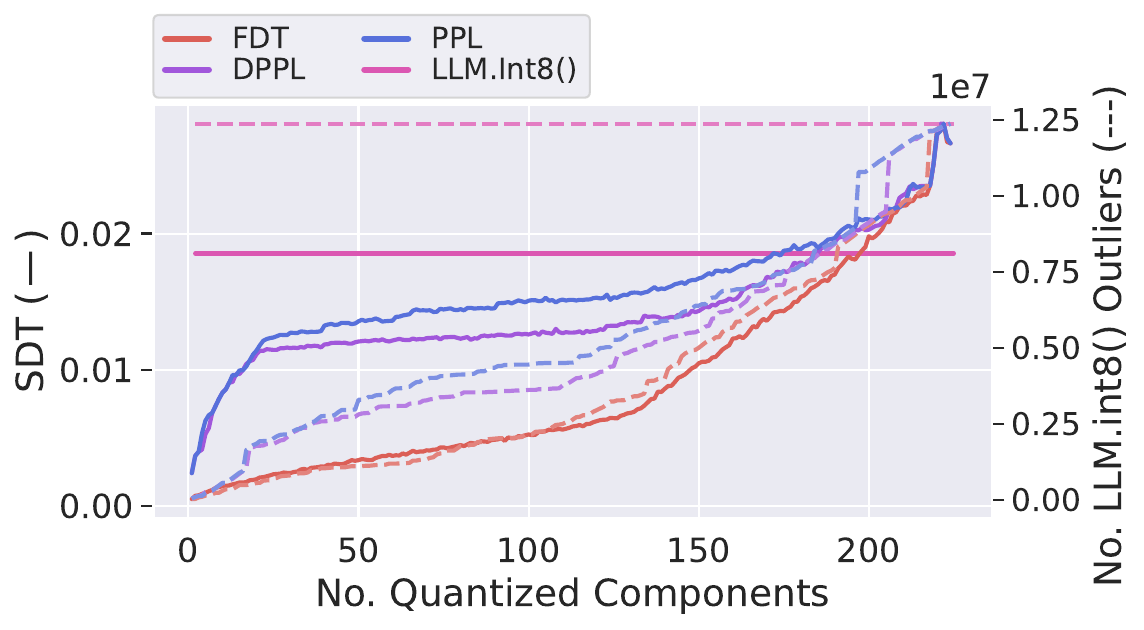}
    \caption{Context size of 25 tokens.}
\end{subfigure}
\begin{subfigure}{0.49\linewidth}
    \centering
    \includegraphics[width=\linewidth]{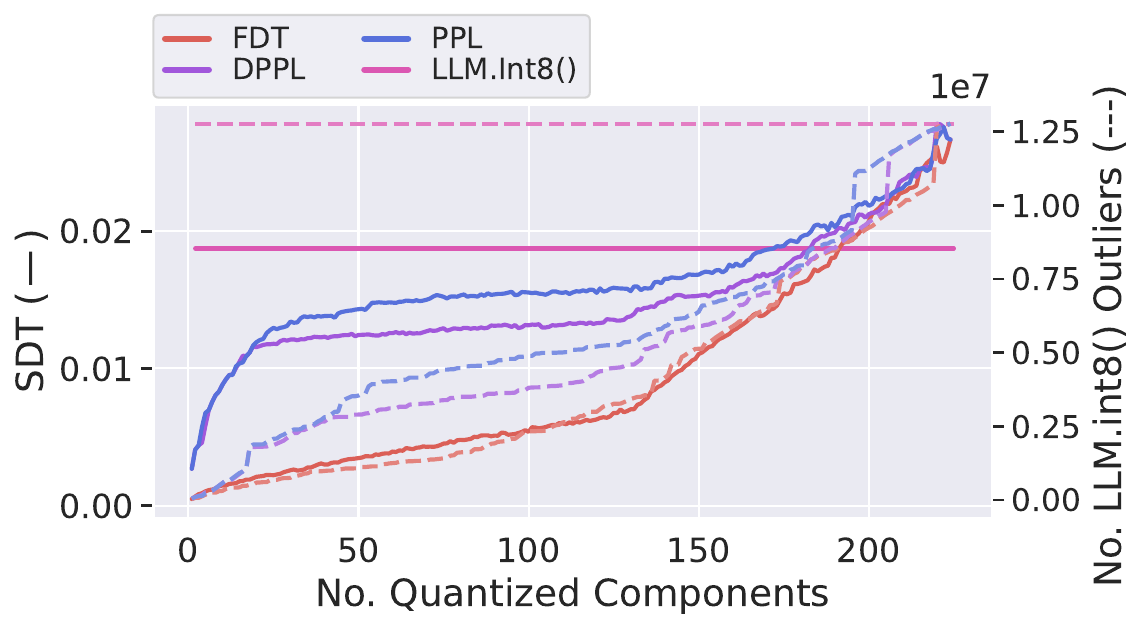}
    \caption{Context size of 50 tokens.}
\end{subfigure}
\begin{subfigure}{0.49\linewidth}
    \centering
    \includegraphics[width=\linewidth]{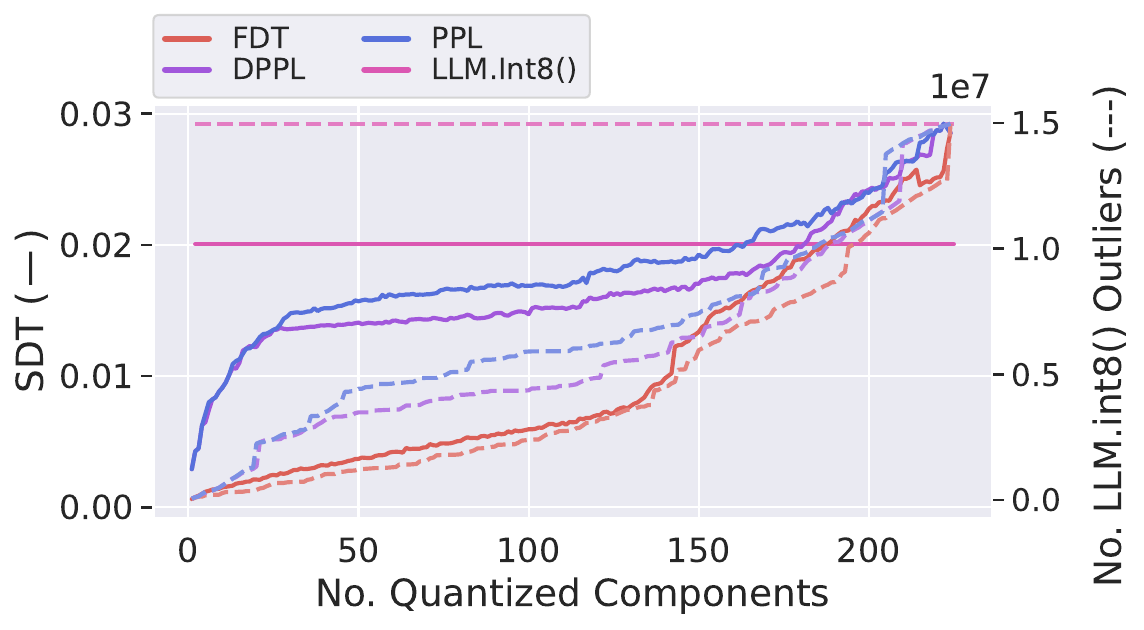}
    \caption{Context size of 200 tokens.}
\end{subfigure}
\begin{subfigure}{0.49\linewidth}
    \centering
    \includegraphics[width=\linewidth]{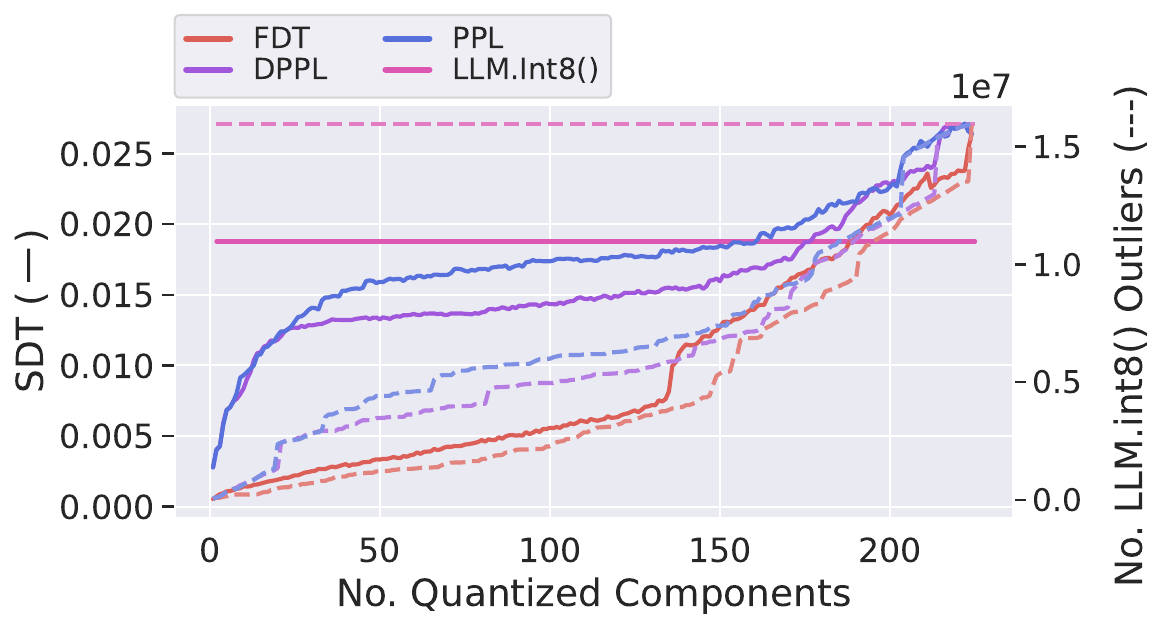}
    \caption{Context size of 300 tokens.}
\end{subfigure}
\begin{subfigure}{0.49\linewidth}
    \centering
    \includegraphics[width=\linewidth]{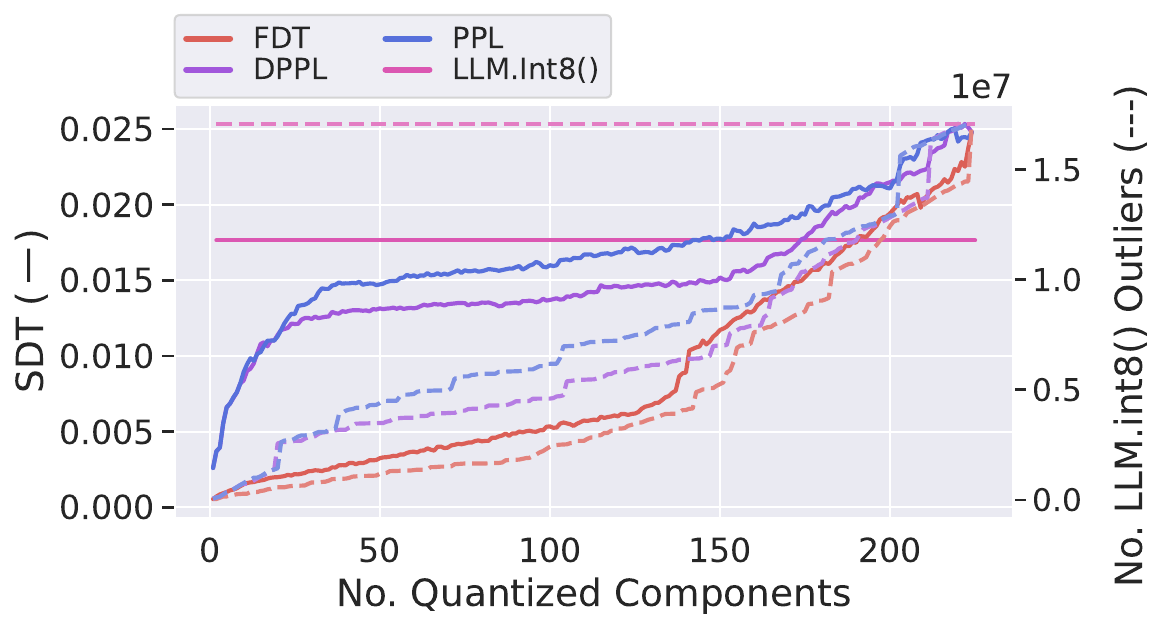}
    \caption{Context size of 400 tokens.}
\end{subfigure}
\caption{Greedy Search Tree results for different context sizes.}
    \label{fig:quant_nlp_greedy}
\end{figure*}

\begin{figure}[t]
    \centering
    \includegraphics[width=\linewidth]{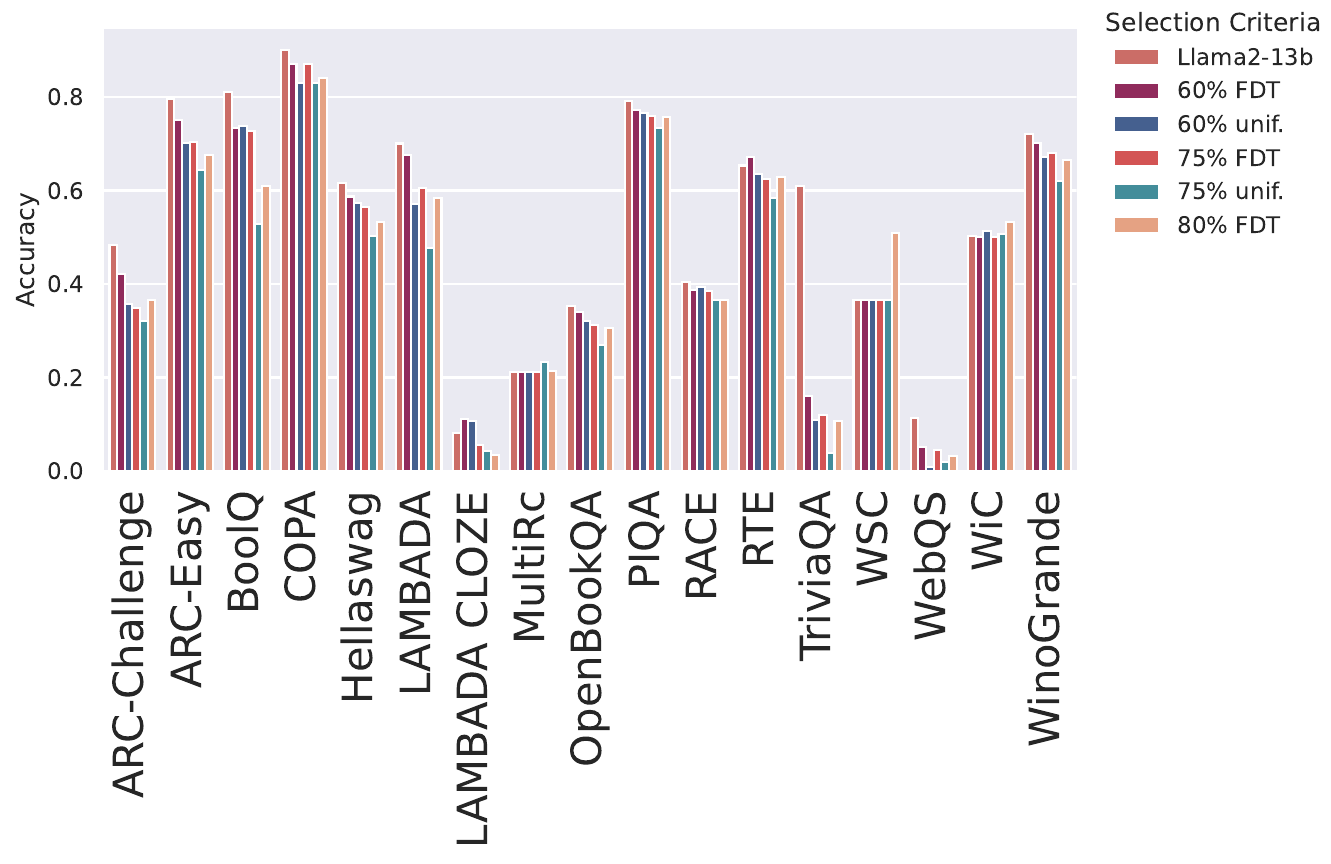}
    \caption{Detailed view on aggregated values of Tab.~\ref{tab:comp_model_benchmarks} when selecting Llama2-13B components to sparsify by metrics.}
    \label{fig:sparse_nlp}
\end{figure}

\begin{figure*}[t]
\begin{subfigure}{0.33\linewidth}
\includegraphics[width=\linewidth]{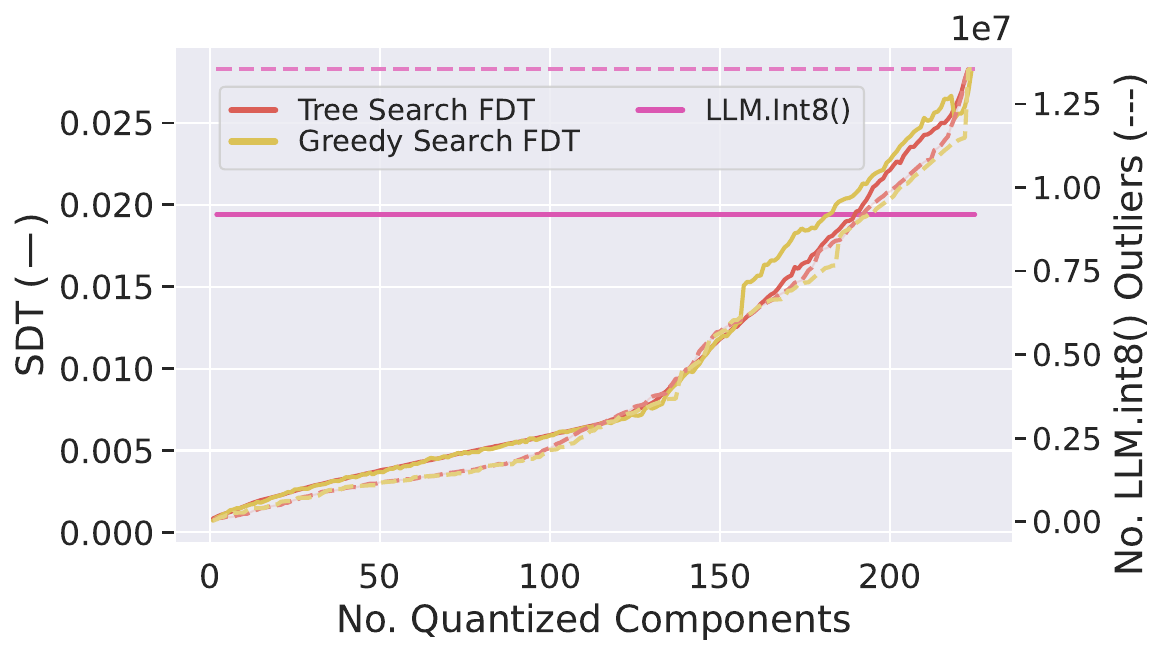}
\caption{FDT tree vs greedy}
\label{fig:comp_fdt}
\end{subfigure}
\begin{subfigure}{0.33\linewidth}
\includegraphics[width=\linewidth]{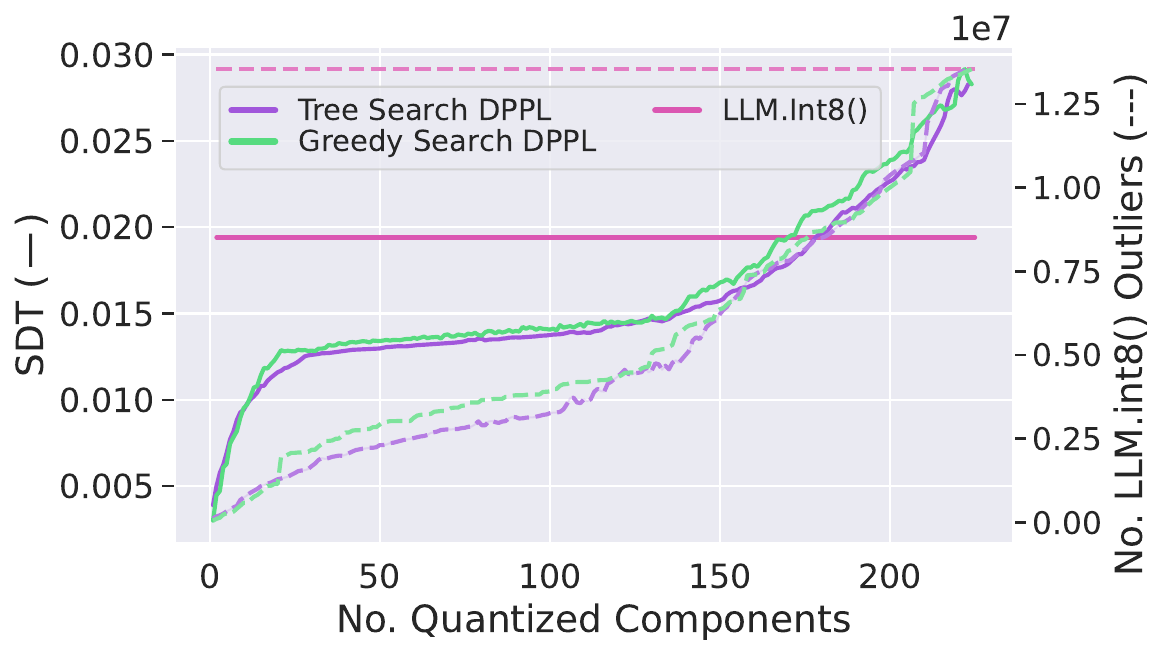}
\caption{DPPL tree vs greedy}
\end{subfigure}
\begin{subfigure}{0.33\linewidth}
\includegraphics[width=\linewidth]{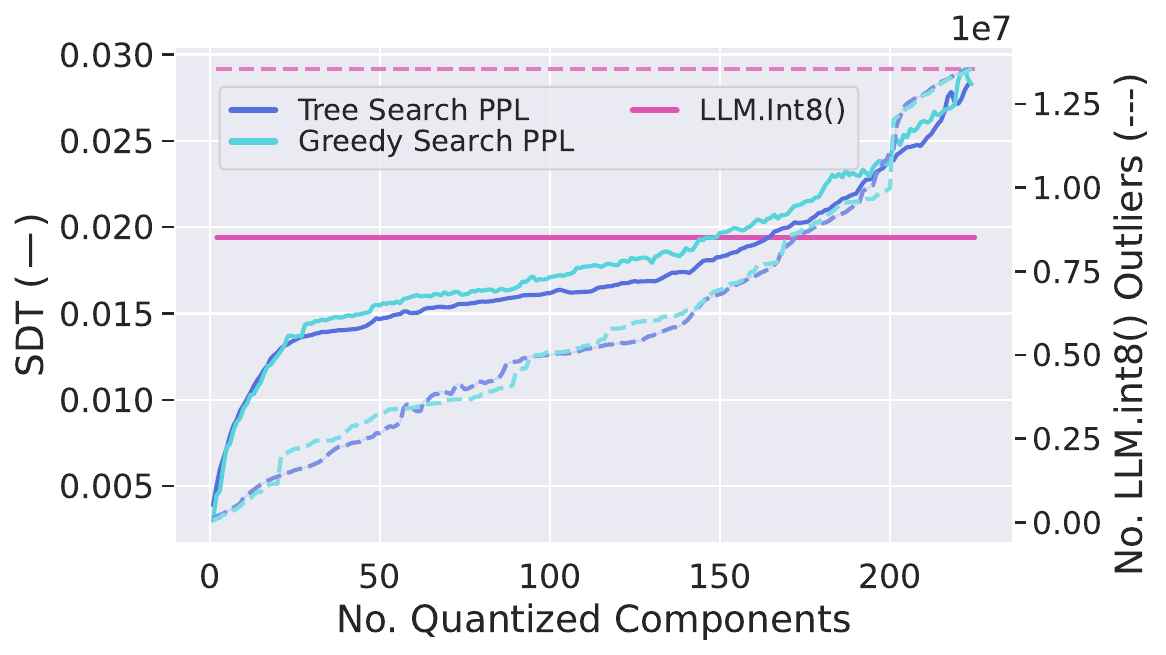}
\caption{PPL tree vs greedy}
\end{subfigure}
\caption{Comparison of performance when selecting components by the tree-search as described to greedy selection of once evaluated components for all discussed metrics. Clearly, FDT is most stable until 150 components.}
\label{fig:comparison_search}
\end{figure*}

\begin{figure*}[t]
\begin{subfigure}{0.48\linewidth}
\includegraphics[width=\linewidth]{images/greedy_tree_comparison_50.pdf}
\caption{Mean sorted greedy tree with 50 context tokens.}
\label{fig:comp_fdt}
\end{subfigure}
\begin{subfigure}{0.48\linewidth}
\includegraphics[width=\linewidth]{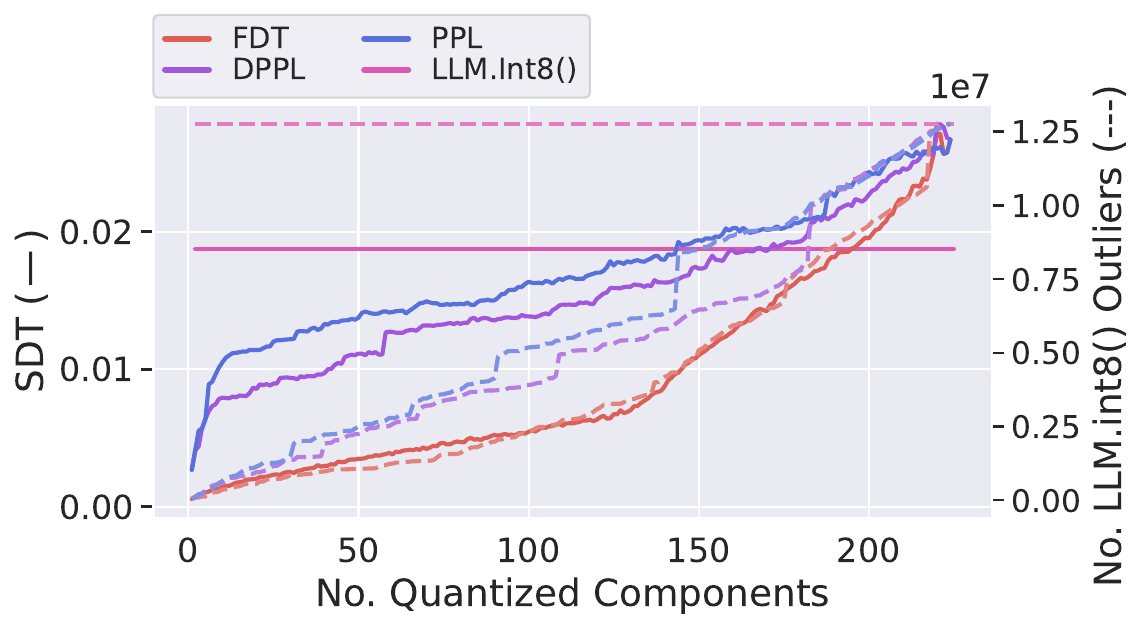}
\caption{Std=0.25 sorted components with 50 context tokens.}
\end{subfigure}

\begin{subfigure}{0.48\linewidth}
\includegraphics[width=\linewidth]{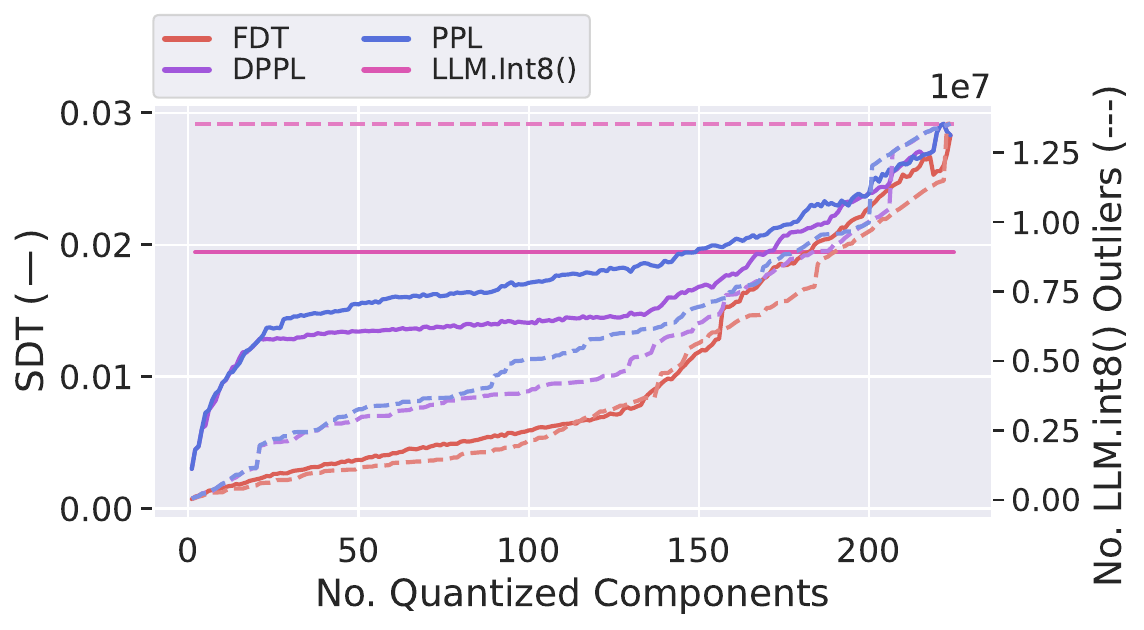}
\caption{Mean sorted greedy tree with 100 context tokens.}
\end{subfigure}
\begin{subfigure}{0.48\linewidth}
\includegraphics[width=\linewidth]{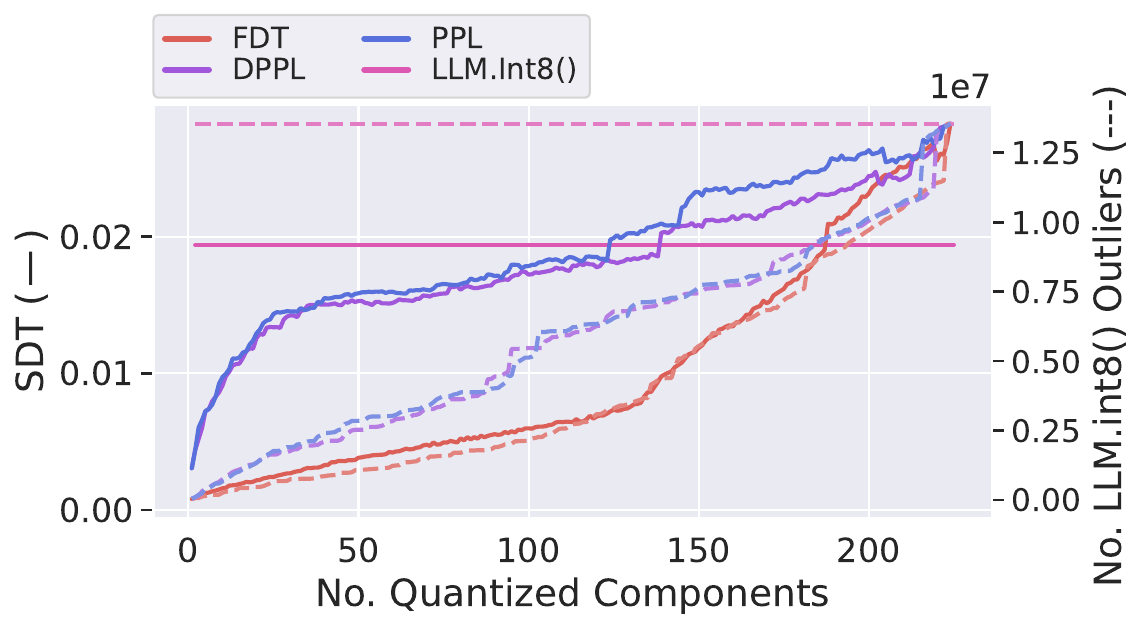}
\caption{Std=0.25 sorted components with 100 context tokens.}
\end{subfigure}
\caption{Comparing the ranking of the components based on mean or standard deviation.}
\label{fig:comparison_search}
\end{figure*}




\section{Details on Quantization Sec.~\ref{sec:quant_tree}}
\begin{figure*}[t]
\centering
    \includegraphics[width=.99\linewidth]{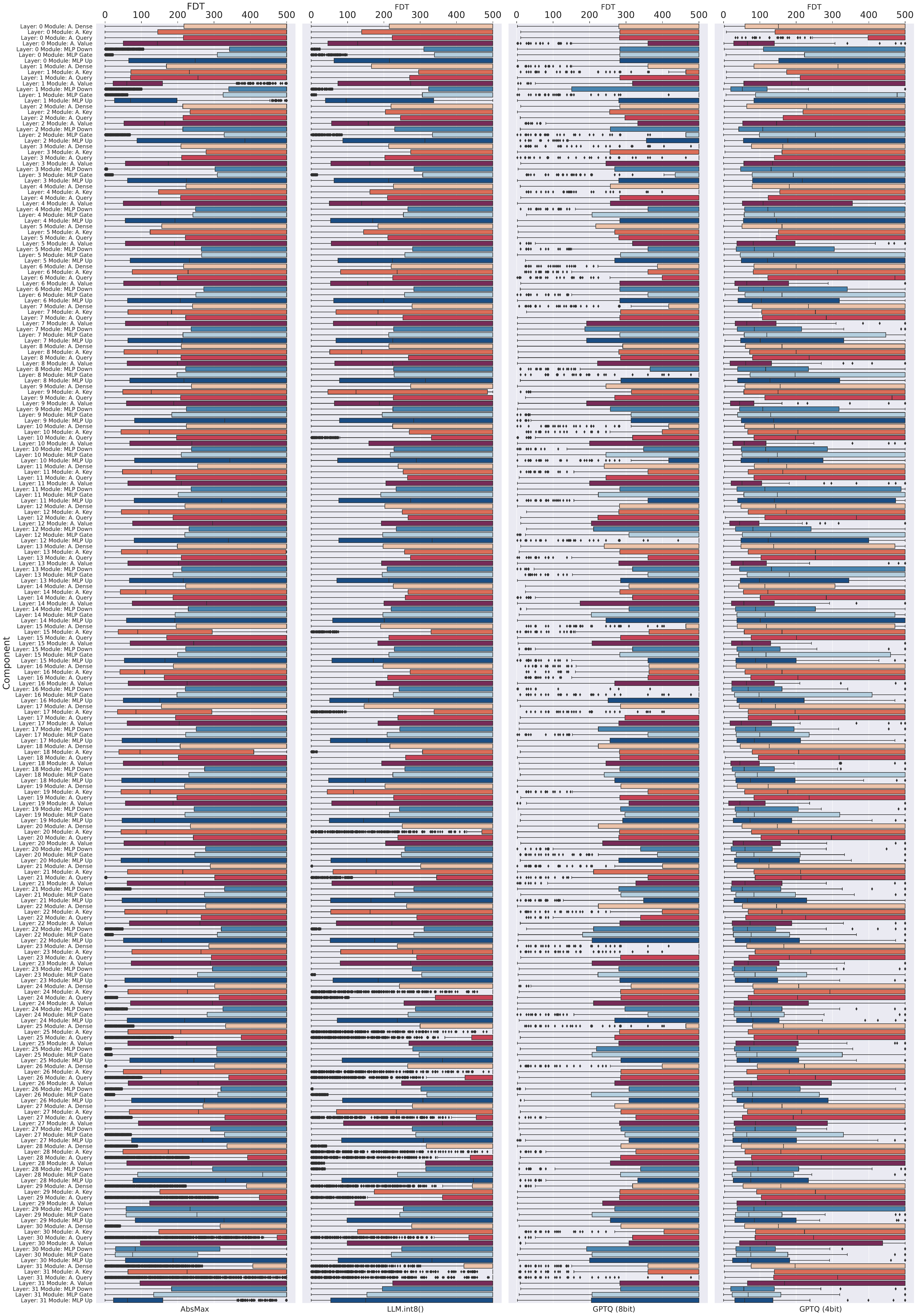}
\caption{Full view of the influence of individual componentwise quantization measured by FDT.}
\label{fig:componentwise_plots}
\end{figure*}

Fig.~\ref{fig:componentwise_plots} shows detailed componentwise evaluations aggregated in Fig.~\ref{fig:comp_quant_violin}.

\begin{figure*}[t]
\begin{subfigure}{\linewidth}
\includegraphics[width=\linewidth]{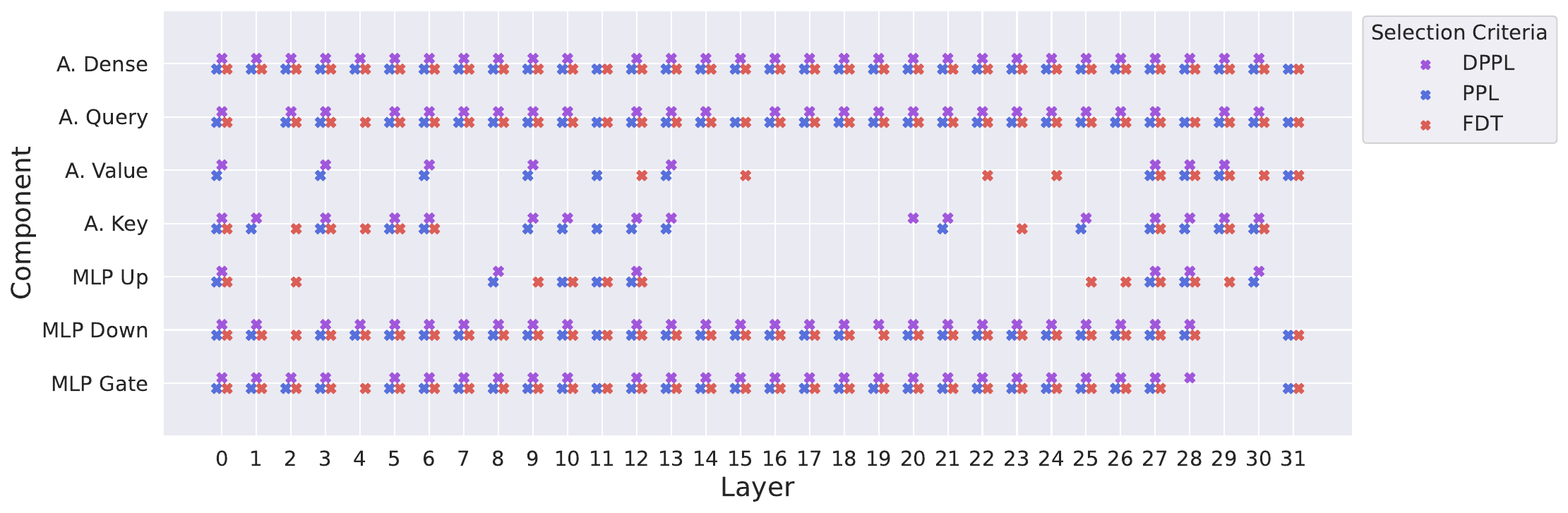}
\caption{The 150 selected components  selected by metrics for 8-bit AbsMax conversion.}
\end{subfigure}
\begin{subfigure}{\linewidth}
\includegraphics[width=\linewidth]{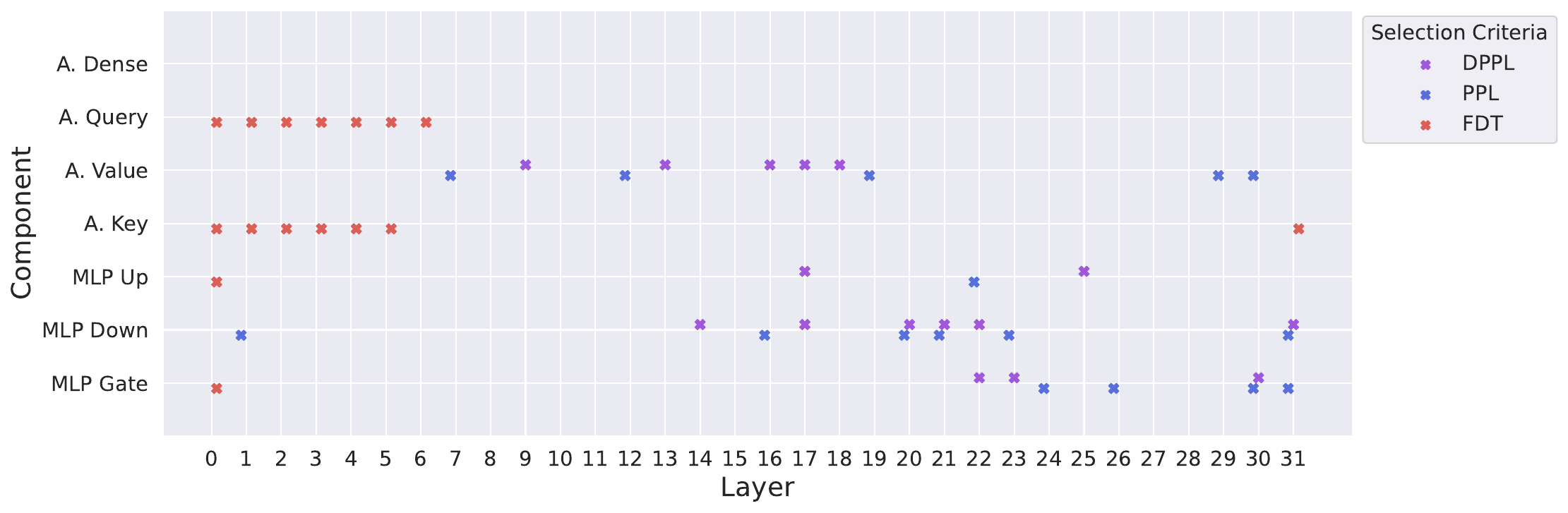}
\caption{The 16 selected components selected by metrics for 4-bit GPTQ conversion.}
\end{subfigure}
\caption{Detailed view of the  Llama2-7B components in Tab.~\ref{tab:comp_model_benchmarks} selected by metrics for lower precision conversion.}
\label{fig:selected_components}
\end{figure*}

Fig.~\ref{fig:selected_components} shows the final configurations as compared in Tab.~\ref{tab:comp_model_benchmarks}.

Fig.~\ref{fig:quant_nlp} shows the detailed nlp-eval scores of Tab.~\ref{tab:comp_model_benchmarks}.

Fig.~\ref{fig:quant_nlp_greedy} shows greedy search trees over various context lengths.

In total the entire search evaluation required 16 GPU-days with A100s to complete all metrics.

\section{Details on Sparsification, Sec.~\ref{sec:sparse}}
\label{app:trend}\label{app:sparse}
\begin{figure}
\begin{subfigure}{\linewidth}
\includegraphics[width=\linewidth]{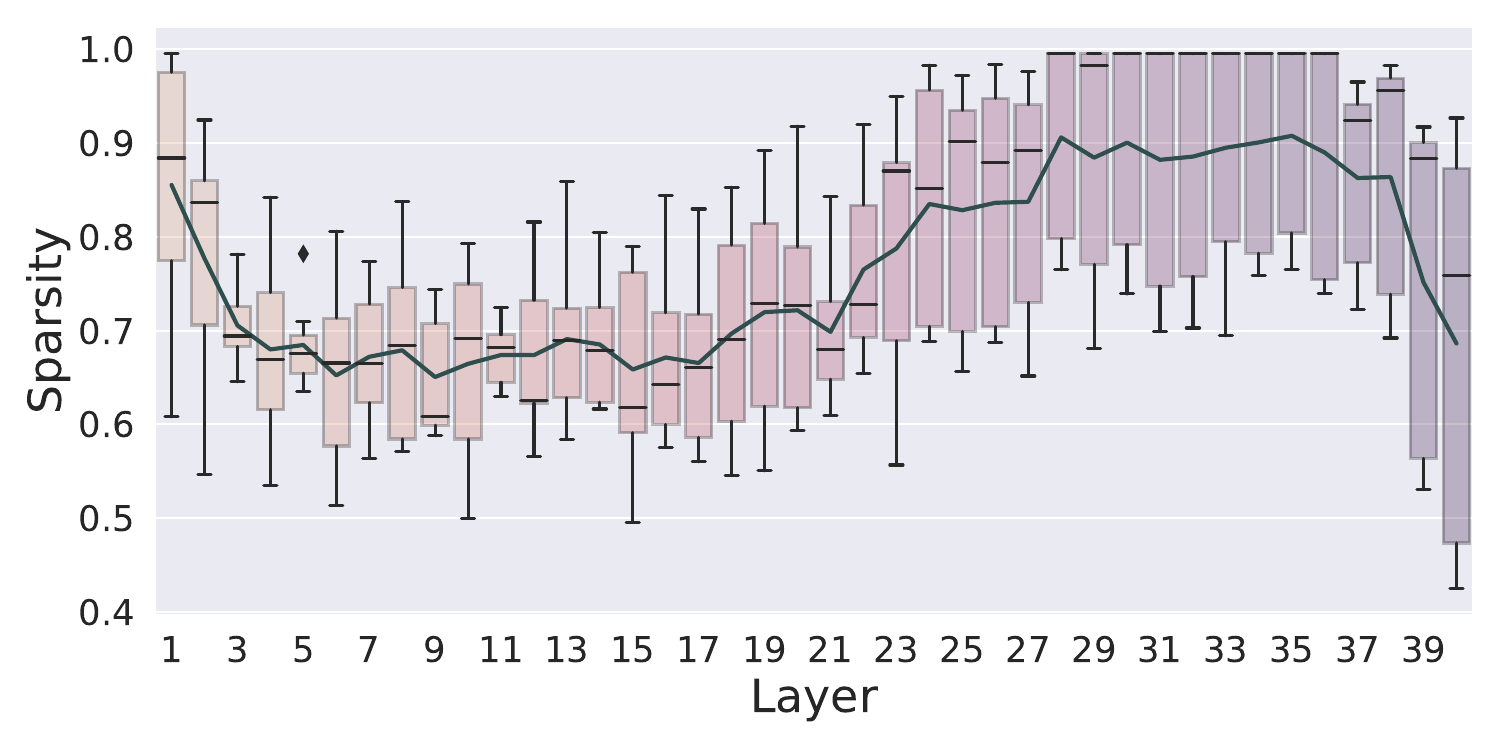}
\end{subfigure}
\begin{subfigure}{\linewidth}
\includegraphics[width=\linewidth]{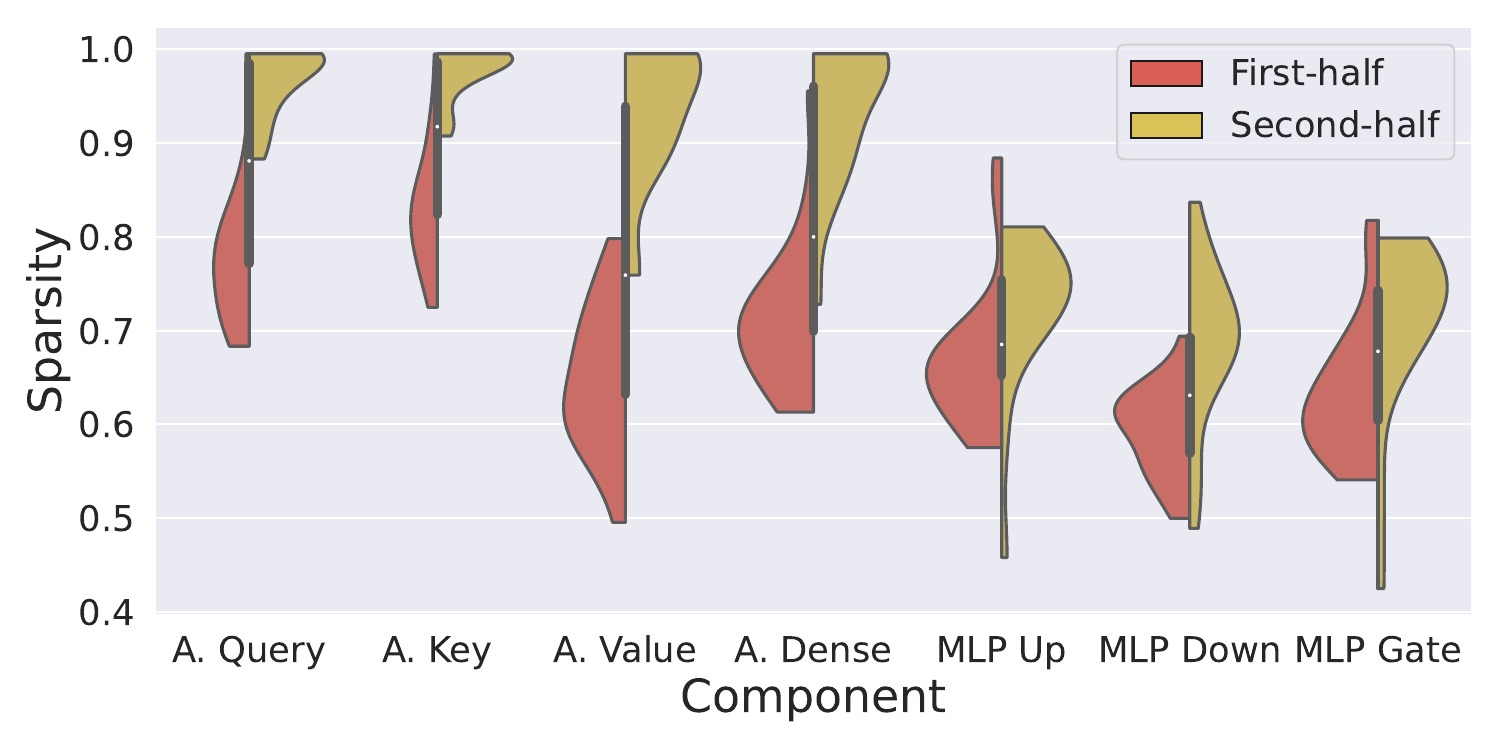}
\end{subfigure}
\caption{Distribution of 75\% average model sparsity. A. denotes Attention.
\textbf{Top}: Aggregated by layers. The first and last layer have highest variance (MLP most important, \textit{cf.} Fig.~\ref{fig:sparse_componentwise}). Second half reaches sparsities close component removal.
\textbf{Bottom}: Per component aggregation. In the second half of layers, the importance of attention drops drastically. MLP almost remains, with outliers to larger importance.}
\label{fig:distribution_sparse}
\end{figure}

Fig.~\ref{fig:distribution_sparse} shows a different aggregated perspective of Fig.~\ref{fig:sparse_componentwise}, to point out more direct the occuring variances.

\begin{figure}[t]
\begin{subfigure}{\linewidth}
    \includegraphics[width=\linewidth]{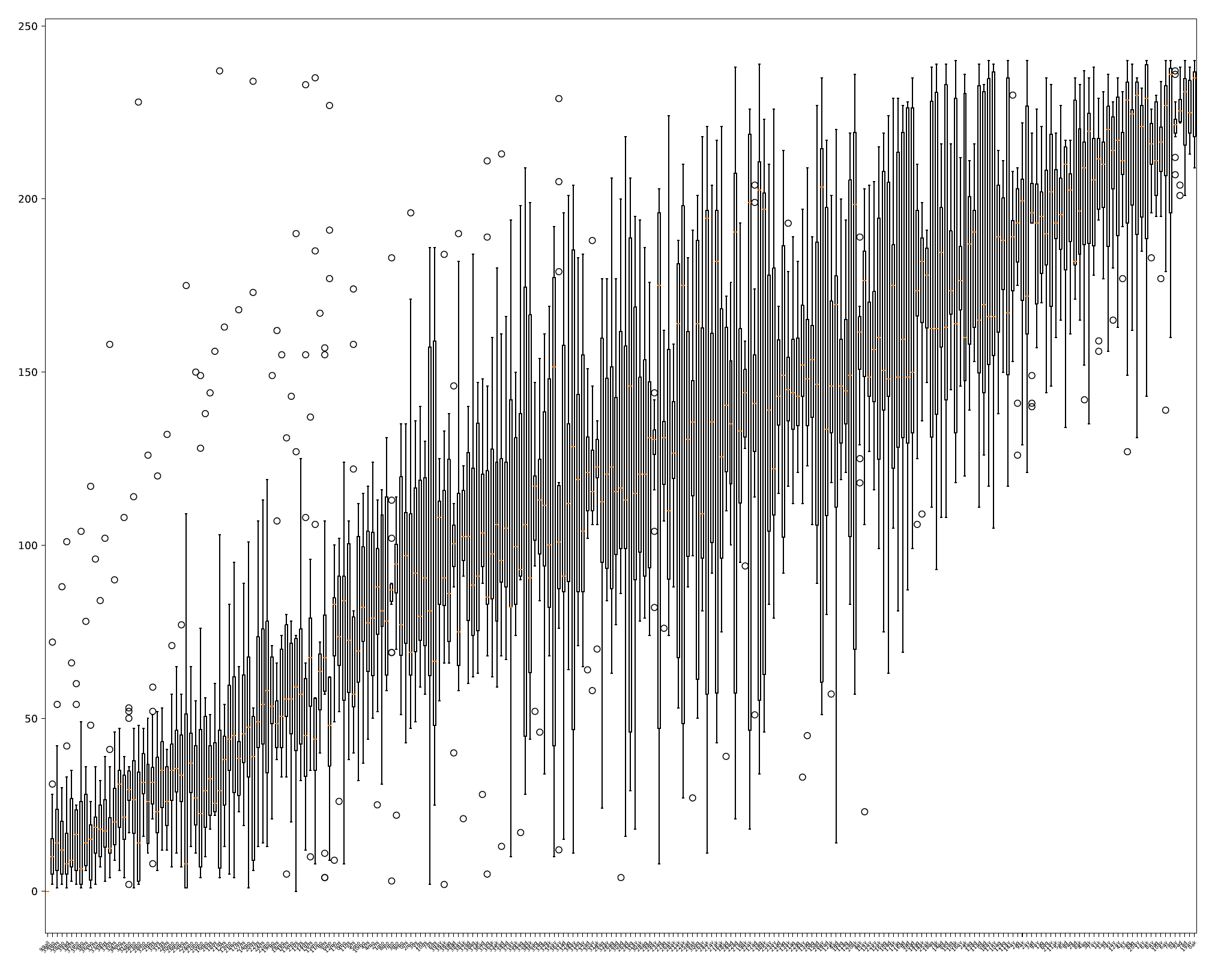}
\end{subfigure}
\caption{Trends during sparsification. We plot the ranking of the components FDT value through various sparsity levels (y-axis) for all components (x-axis). Interestingly, there is a clear trend of components retaining ``their importance''.}
\label{fig:evidence}\label{fig:trend}
    \label{fig:sparse_trend}
\end{figure}

Fig.~\ref{fig:sparse_trend} shows the rank of lowest influence (measured by FDT) of components (x-axis) throughout various sparsity levels (y-axis). I.e. starting with a uniformly pruned model in 5\% steps, we measured the rank when adding an additional 2.5\% only to a single component.
Interestingly, components seem to retain their importance throughout the various levels of sparsity. 

Fig.~\ref{fig:sparse_nlp} shows the detailed nlp-eval scores of Tab.~\ref{tab:comp_model_benchmarks}.

Note that, despite being often close in relative sparsity, the total number of parameters pruned for MLP is significantly larger than for Attention matrices (ratio 3:1).

\begin{figure}[t]
\vspace{-2cm}
\begin{subfigure}{\linewidth}
    \includegraphics[angle=-90,width=.35\linewidth]{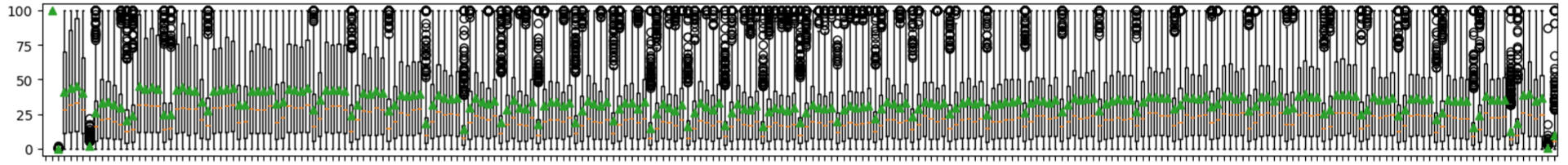}
\end{subfigure}
\caption{In style of Fig.~\ref{fig:componentwise_plots}, Boxplots of the FDT value for all components, when pruning the lowest 60\% weights of a single at once for Llama-13B.
Almost depicting a sinus curve, ``most equal'' relevance of components appear in the middle of the model. The first and last layers are dominated by outliers, i.p., MLP Up and A. Dense.}
\label{fig:prune_component}
\end{figure}

Finally, Fig.~\ref{fig:prune_component} shows detailed componentwise pruning experiments.

In total one sparsification training required 32 GPU-days with A100s for our experiment, and 29 GPU-days for uniform pruning.

\section{Further Hyperparameters}
\begin{figure}[t]
\centering
    \includegraphics[width=.3\linewidth]{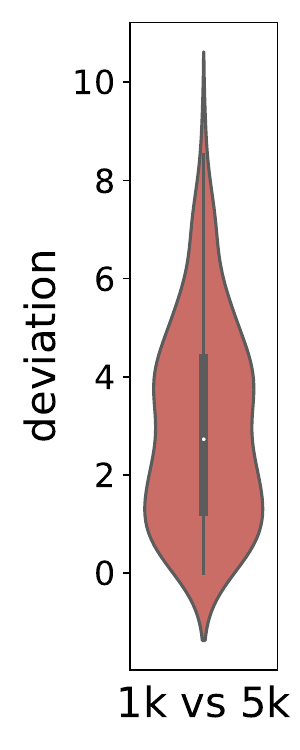}
    \caption{Boxplot of the absolute differences in $\mathcal{M}_{F_{75}}$ value  for 1000 experiments comparing aggregating with 1k and 5k samples of $F_{75}$ values.}
    \label{fig:boxplot_1k5k}
\end{figure}

Fig.~\ref{fig:boxplot_1k5k} shows that 1k probes are enough to determine a stable $\mathcal{M}_{F_{75}}$ value.

\section{Model Completion vs. Ground Truth}
\label{sec:compl_gt}

\begin{figure}[t]
\centering
    \includegraphics[width=.8\linewidth]{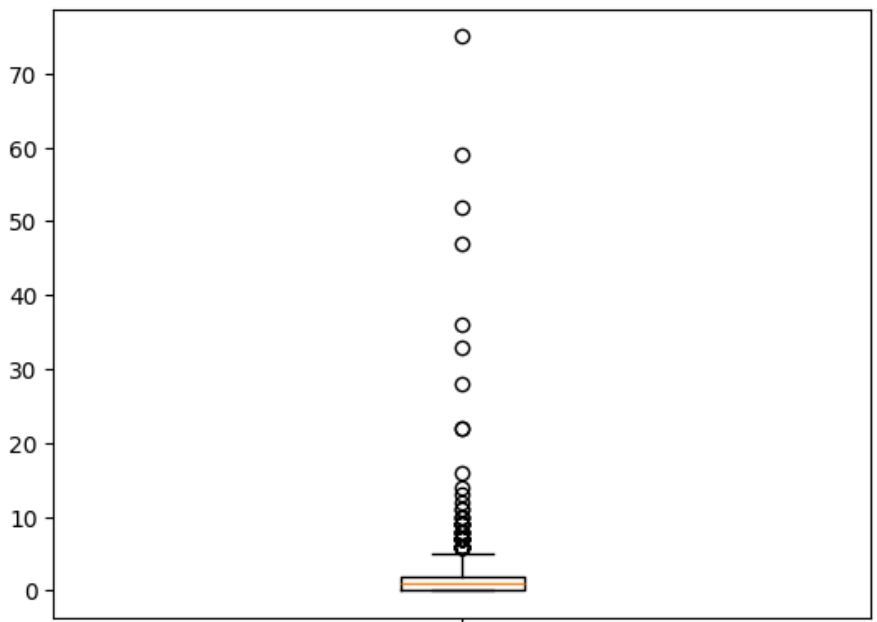}
    \caption{Boxplot shows FDT value (y-axis) for 1000 samples of the original Wikitext2 completions and LLama2-13b completions.}
    \label{fig:boxplot_fdt_original_vs_completion}
\end{figure}

Fig.~\ref{fig:boxplot_fdt_original_vs_completion} shows the FDT value that compares the ground truth of the data set with the completion of the model on Wikitext2. Tab.~\ref{tab:my_label} further gives an example for better comparison. It can be observed that, even though Wikitext2 was contained in the training dataset, Llama2-13B does not properly recall it, and as such, the FDT value is close to 0. Therefore, we cannot utilize generic data samples as ground truth alone in such a general-purpose pruning setup. Afterall, our goal is to measure the ``deviation'' of the compressed model to its baseline. For more specific, e.g. Q/A datasets, it may, however, be more feasible.
\newpage

\newpage
\begin{table*}
    \centering
    \begin{tabular}{p{\linewidth}}
\textbf{Prefix:}\\
`Where On June 19 , 1865 , Union General Gordon Granger , standing on its balcony , read the contents of “ General Order No. 3 ” , thereby emancipating all slaves in the state of Texas . St. Joseph ’ s Church was built by German immigrants in 1859 @-@ 60 and is the oldest wooden church building in Galveston and the oldest German Catholic Church in Texas . The church was '\\
\textbf{Ground-truth:}\\
`dedicated in April 1860 , to St. Joseph , the patron saint of laborers . The building is a wooden gothic revival structure , rectangular with a square bell tower with trefoil window . The U.S. Custom House began construction in 1860 and was completed in 1861 . The Confederate Army occupied the building during the American Civil War , In 1865 , the Custom House was the site of the'\\
\textbf{Llama2-7B completion:}\\
`built in the Gothic Revival style , and its bell tower has a spire that rises 125 feet . The church is located in the East End Historic District , which is listed on the National Register of Historic Places . The church was added to the National Register of Historic Places in 1975 . The Strand National Historic Landmark District is a National Historic Landmark District in Galveston , Texas , United States . It is located on the Strand ,'\\
\textbf{LLM.int8(Llama2-7B) completion:}\\
`originally located on 25th Street between Broadway and Strand Street in Galveston . It was moved to its present location in 1892 . The church was designated a Recorded Texas Historic Landmark in 1964 , listed on the National Register of Historic Places in 1975 , and designated a National Historic Landmark in 2000 . The Galveston Daily News Building is a historic commercial building in Galveston , Texas .'
    \end{tabular}
    \caption{Example of a Wikitext2 probe as used to determined the FDT value. 
    It can be observed that the fully LLM.int8 compressed model, though having only minor differences in the NLP and PPL benchmarks, substantially alters the original model completion, starting at the very first token.}
    \label{tab:my_label}
\end{table*}


\end{document}